\documentclass[nohyperref]{article}

\usepackage{microtype}
\usepackage{graphicx}
\usepackage{subfigure}
\usepackage{booktabs} % for professional tables
\usepackage{listings} % for code snippets
\usepackage[ragged]{footmisc}

\usepackage{hyperref}

\usepackage[accepted]{icml2022}
\usepackage{amsmath}
\usepackage{amssymb}
\usepackage{mathtools}
\usepackage{amsthm}
\usepackage{bm}

\usepackage{url}
\usepackage{tabularx}
\usepackage{multirow}
\usepackage{colortbl}
\usepackage{glossaries}

\usepackage[capitalize,noabbrev, nameinlink]{cleveref}

\theoremstyle{plain}

\theoremstyle{definition}

\theoremstyle{remark}

\usepackage[textsize=tiny]{todonotes}

\icmltitlerunning{From data to functa}

\begin{document}

\twocolumn[
\icmltitle{From data to functa: Your data point is a function \\ and you can treat it like one}

\icmlsetsymbol{equal}{*}

\begin{icmlauthorlist}
\icmlauthor{Emilien Dupont}{equal,oxford}
\icmlauthor{Hyunjik Kim}{equal,dm}
\icmlauthor{S. M. Ali Eslami}{dm}
\icmlauthor{Danilo Rezende}{dm}
\icmlauthor{Dan Rosenbaum}{haifa,dm}
\end{icmlauthorlist}

\icmlaffiliation{oxford}{University of Oxford}  % Department of Statistics, 
\icmlaffiliation{haifa}{University of Haifa}  % Department of Computer Science, 
\icmlaffiliation{dm}{DeepMind}

\icmlcorrespondingauthor{Emilien Dupont}{dupont@stats.ox.ac.uk}
\icmlcorrespondingauthor{Hyunjik Kim}{hyunjikk@google.com}

\icmlkeywords{Machine Learning, ICML}

\vskip 0.3in
]

\printAffiliationsAndNotice{\icmlEqualContribution}

\begin{abstract}
It is common practice in deep learning to represent a measurement of the world on a discrete grid, e.g.\ a 2D grid of pixels. However, the underlying signal represented by these measurements is often continuous, e.g.\ the scene depicted in an image. A powerful continuous alternative is then to represent these measurements using an \textit{implicit neural representation}, a neural function trained to output the appropriate measurement value for any input spatial location. In this paper, we take this idea to its next level: what would it take to perform deep learning on these functions instead, treating them as data? In this context we refer to the data as \textit{functa}, and propose a framework for deep learning on functa. This view presents a number of challenges around efficient conversion from data to functa, compact representation of functa, and effectively solving downstream tasks on functa. We outline a recipe to overcome these challenges and apply it to a wide range of data modalities including images, 3D shapes, neural radiance fields (NeRF) and data on manifolds. We demonstrate that this approach has various compelling properties across data modalities, in particular on the canonical tasks of generative modeling, data imputation, novel view synthesis and classification. Code: \href{https://github.com/deepmind/functa}{\nolinkurl{github.com/deepmind/functa}}.
\end{abstract}

\section{Introduction}

In deep learning, data is traditionally represented by arrays. For example, images are represented by their pixel intensities, and 3D shapes by voxel occupancies, both at a discrete set of grid coordinates tied to a particular resolution. However, the underlying signal represented by these arrays is often continuous. It is therefore natural to consider representing such data with continuous quantities.

Recently, the idea of modelling data with continuous functions has gained popularity. An image, for example, can be represented by a continuous function mapping 2D pixel coordinates to RGB values. When such a function is parameterized by a neural network, it is typically referred to as an \textit{implicit neural representation} (INR). INRs are generally applicable to a wide range of modalities -- indeed, various works have demonstrated that INRs can be used to represent images \cite{stanley2007compositional, ha2016generating}, 3D shapes \cite{mescheder2019occupancy, chen2019learning}, signed distance functions \cite{park2019deepsdf}, videos \cite{li2021neural}, 3D scenes \cite{mildenhall2020nerf}, audio \cite{sitzmann2020implicit} and data on manifolds \cite{dupont2021generative}. This functional representation offers a number of advantages over array representations. It allows for dealing with data at arbitrary resolutions, as well as data that is difficult to discretize such as neural radiance fields (NeRF) for 3D scene representation \cite{mildenhall2020nerf}. Parameterizing such functions as neural networks offers additional advantages, in terms of memory-efficiency and as a single architecture that can represent different data modalities.

\begin{figure}[t]
\begin{center}
\includegraphics[width=\columnwidth]{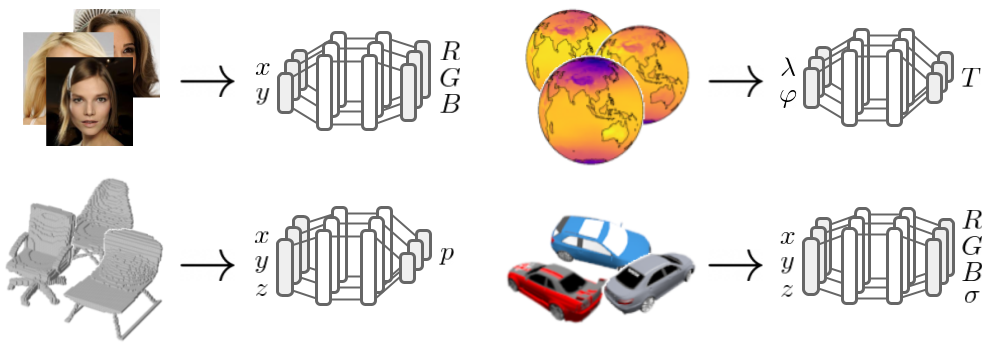}

\vspace{7pt}

\includegraphics[width=\columnwidth]{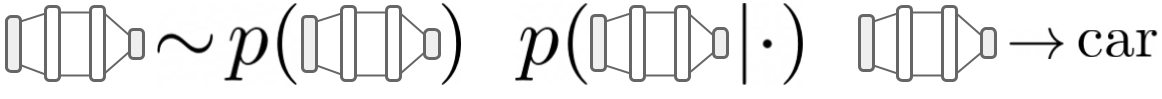}
\end{center}
\vspace{-5pt}
\hspace{3pt} Generative modeling \hspace{20pt} Inference \hspace{20pt} Classification 
\caption{We convert array data into functional data parameterized by neural networks, termed \textit{functa}, and treat these as data points for various downstream machine learning tasks.}
\vspace{-3mm}
\label{fig:figure1}
\end{figure}

In light of these advantages, we propose a new framework that 1.\ converts array data to functional data parameterized by neural networks, and 2.\ performs deep learning tasks directly on these functions. The first step involves taking a dataset of a given modality and fitting an INR to each datapoint. We refer to these functions as \textit{functa}, a concise term for \textit{INRs that are to be thought of as data}. In the second step, we treat functa as data points that we use for deep learning tasks (see \autoref{fig:figure1}). A key difference to prior work on multimodal learning of functions \cite{dupont2021generative, du2021gem} is that we decouple the creation of datasets of functa in the first step and the deep learning task in the second step (e.g.\ generative modeling, inference, classification). Functions are then treated as data rather than part of the model for solving the task at hand.

While this framework inherits the advantages of INRs, we are also presented with new challenges. Which parameterization of functa do we choose? How do we efficiently create large datasets of functa, or functasets? How can we use functa as inputs to neural networks for downstream deep learning tasks? We are therefore required to lay the groundwork for methodology that is appropriate for working with data points as functions. In this paper we propose an instantiation of this framework, in particular a unified method for efficiently creating large functasets for a wide range of data modalities, including images, voxels, NeRF scenes and data on manifolds. We choose a specific parameterization of functa, called \textit{modulations}, that can be fed into neural architectures to solve various downstream tasks including generative modeling, data imputation, novel view synthesis and classification. While our approach does not yet outperform conventional deep learning on arrays, our results show that the functional view has several desirable properties and is a promising alternative to traditional data representation.

\section{Functa: Data points as INRs} \label{sec:functa}

We first review INRs then discuss the advantages of using them as data points (functa), as well as the advantages of decoupling the conversion of data to functa and the downstream deep learning task. INRs are functions $f_\theta: \mathcal{X} \to \mathcal{F}$ mapping \textit{coordinates} $\mathbf{x} \in \mathcal{X}$ (e.g.\ pixel locations) to \textit{features} $\mathbf{f} \in \mathcal{F}$ (e.g.\ RGB values) with parameters $\theta$. Given a data point as a collection of coordinates $\{ \mathbf{x}_i \}_{i \in \mathcal{I}}$ and features $\{ \mathbf{f}_i \}_{i \in \mathcal{I}}$ (where $\mathcal{I}$ is an index set corresponding to e.g.\ all pixel locations in an image), INRs are fitted by minimizing mean squared error over all coordinate locations:
\begin{equation} \label{eq:inr-opt}
    \min_{\theta} \mathcal{L}(f_{\theta}, \{ \mathbf{x}_i, \mathbf{f}_i\}_{i \in \mathcal{I}}) = \min_{\theta} \sum_{i \in \mathcal{I}} \| f_\theta(\mathbf{x}_i) - \mathbf{f}_i \|^2_2.
\end{equation}
Hence each $f_\theta$ corresponds to e.g. a single image. Typically, $f_\theta$ is parameterized by a feedforward neural network (MLP) with positional encodings \cite{mildenhall2020nerf, tancik2020fourier} or sinusoidal activation functions \cite{sitzmann2020implicit} that allow fitting of high frequency signals.

\begin{figure}[t]
\begin{center}
\includegraphics[width=\columnwidth]{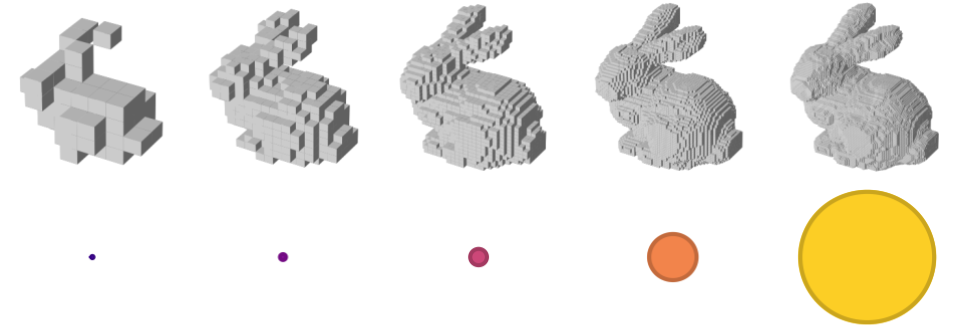}
\end{center}

\vspace{-5pt}

\hspace{13pt} \small{$512$} \hspace{31pt} \small{$4\text{k}$} \hspace{32pt} \small{$32\text{k}$} \hspace{28pt} \small{$262\text{k}$} \hspace{26pt} \small{$2\text{m}$} 

\vspace{-5pt}

\begin{center}
\noindent\rule{0.9\columnwidth}{0.3pt}
\end{center}

\vspace{-7pt}

\begin{center}
\includegraphics[width=\columnwidth]{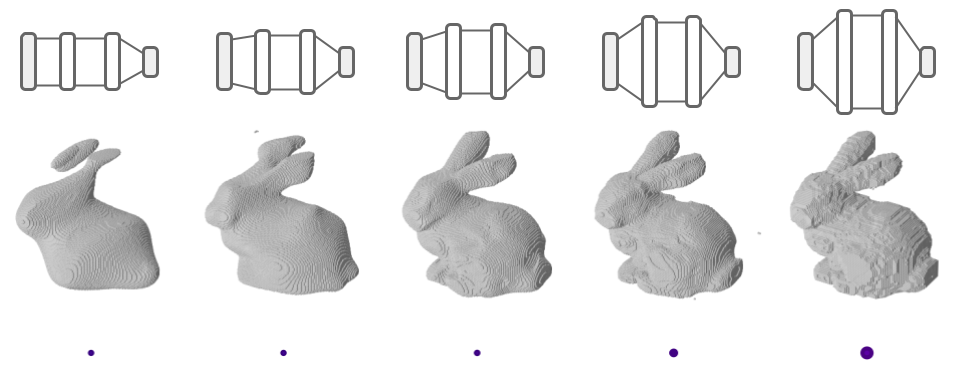}
\end{center}

\vspace{-5pt}

\hspace{12pt} \small{$157$} \hspace{30pt} \small{$257$} \hspace{29pt} \small{$685$} \hspace{29pt} \small{$2.6\text{k}$} \hspace{26pt} \small{$9.6\text{k}$} 

\vspace{-4pt}
\caption{Functa scale much more gracefully with resolution than array representations. Circle area reflects the numerical size of the array (top) / function (bottom). See \Cref{sec:figure2} for details.}
\label{fig:curse-of-disc}
\vspace{-3mm}
\end{figure}

When considered as elements of a dataset, rather than a model, we refer to INRs as \textit{functa}. Training deep learning models on functasets rather than conventional array datasets has various compelling properties that we describe below.

\textbf{Scaling}. Array based representations typically scale poorly with resolution (e.g.\ voxel grids scale cubically with resolution). Processing such high resolution data with neural networks is both memory and compute intensive, often becoming a bottleneck \cite{park2019deepsdf, mescheder2019occupancy}. In contrast, functa usually scale much more gracefully with resolution (\autoref{fig:curse-of-disc}), leading to more efficient training of neural networks for downstream tasks.
% As shown later, high dimensional but low entropy datasets are particularly amenable to being represented as functasets, precisely because they are very compressible. 

\textbf{Moving away from fixed resolution}. Array representations of data are typically stored at a fixed resolution. However, most data in the wild do not follow this form: images come in various shapes and resolutions; data modalities such as lidar are stored as point clouds, 3D shapes as irregular meshes, etc. On the other hand, we can easily convert data at a variety of resolutions to functa, allowing the downstream model to handle data at a range of resolutions.
% \dan{maybe emphasize a direct benefit here: models trained on a specific resolution are not transferable to other resolutions, in contrast models trained on functa can be expected to be agnostic to the resolution.} 

\textbf{Signals that are inherently difficult to discretize}. While visual signals are continuous, they are easily discretized onto a grid to generate digital images. However, this is not true for many other data modalities which are inherently difficult to discretize. For example, neural radiance fields (NeRF) use volumetric rendering which requires the field to be evaluated at arbitrary spatial locations \cite{mildenhall2020nerf} and physical fields often require continuous derivatives to solve differential equations of motion, both of which are incompatible with discrete representations. In addition, it is non-trivial to choose a grid for discretizing data lying on manifolds. Functa provide a natural way to express data continuously and as such bypass the need for discretization when training models on downstream tasks.

\textbf{Multimodality}. It is standard practice in deep learning to use specialized encoder and decoder architectures for different data modalities. However designing encoders for NeRF scenes, for example, is challenging and requires aggregating a collection of images and poses \cite{kosiorek2021nerf}. Similarly, data lying on manifolds require highly specialized architectures \cite{cohen2018spherical}. In contrast, functa can be used to encode and decode a wide variety of data modalities through a generic optimization procedure.

\textbf{Easing downstream task}. The decoupling of 1. creating functa and 2. training models on functa for a downstream task can greatly simplify the task. For example, consider the task of generative modelling on NeRF scenes. Fitting NeRF scenes requires inferring 3D structure from a set of posed 2D images, itself a non-trivial task. Without the above decoupling, learning a generative model of NeRF scenes then requires simultaneously learning a distribution over 3D scenes while inferring 3D structure from 2D images. In our framework, we first infer the 3D scene from 2D images by minimising reconstruction error on views - a relatively easier task - and only after we obtain a functaset of 3D scenes, we model their distribution.

\section{Functasets: Datasets of INRs}

We now describe how to 1.\ represent functa suitably so that they can be fed into neural networks for downstream tasks and 2.\ create large datasets of functa in a scalable manner.

\subsection{Functa as MLP modulations} \label{sec:modulations}

We use SIREN \cite{sitzmann2020implicit} as the base architecture for INRs throughout, since it is known to efficiently represent a wide range of data modalities (\textit{cf.}\ \Cref{sec:siren-modulation} for mathematical formulation of SIREN). The na{\"i}ve approach for representing functa is to take the parameter vector of the SIREN. However, as SIREN is an MLP, it can have a large number of parameters despite the favourable scaling characteristics compared to array representations, making this representation suboptimal for feeding into neural networks for downstream tasks. Various works have explored \textit{modulations} as an alternative that uses a shared base network across data points to model common structure, with modulations modeling the variation specific to each data point \cite{perez2018film, chan2020pi, mehta2021modulated}. We therefore work with modulations rather than parameters, which are typically much more low dimensional.

Modulations are usually represented as elementwise affine transformations (shift and scale) applied to the activations of the neural network \cite{perez2018film, chan2020pi, mehta2021modulated}. However, we found experimentally that using shifts only performs just as well as using both shifts \& scales, with half the representation size. 
We provide the mathematical formulation of \textit{shift modulations} in \Cref{sec:siren-modulation}, along with intuition for the role of shift modulations via perturbation analysis in \Cref{sec:perturbation-analysis}.

\begin{figure}[t]
\begin{center}
\includegraphics[width=\columnwidth]{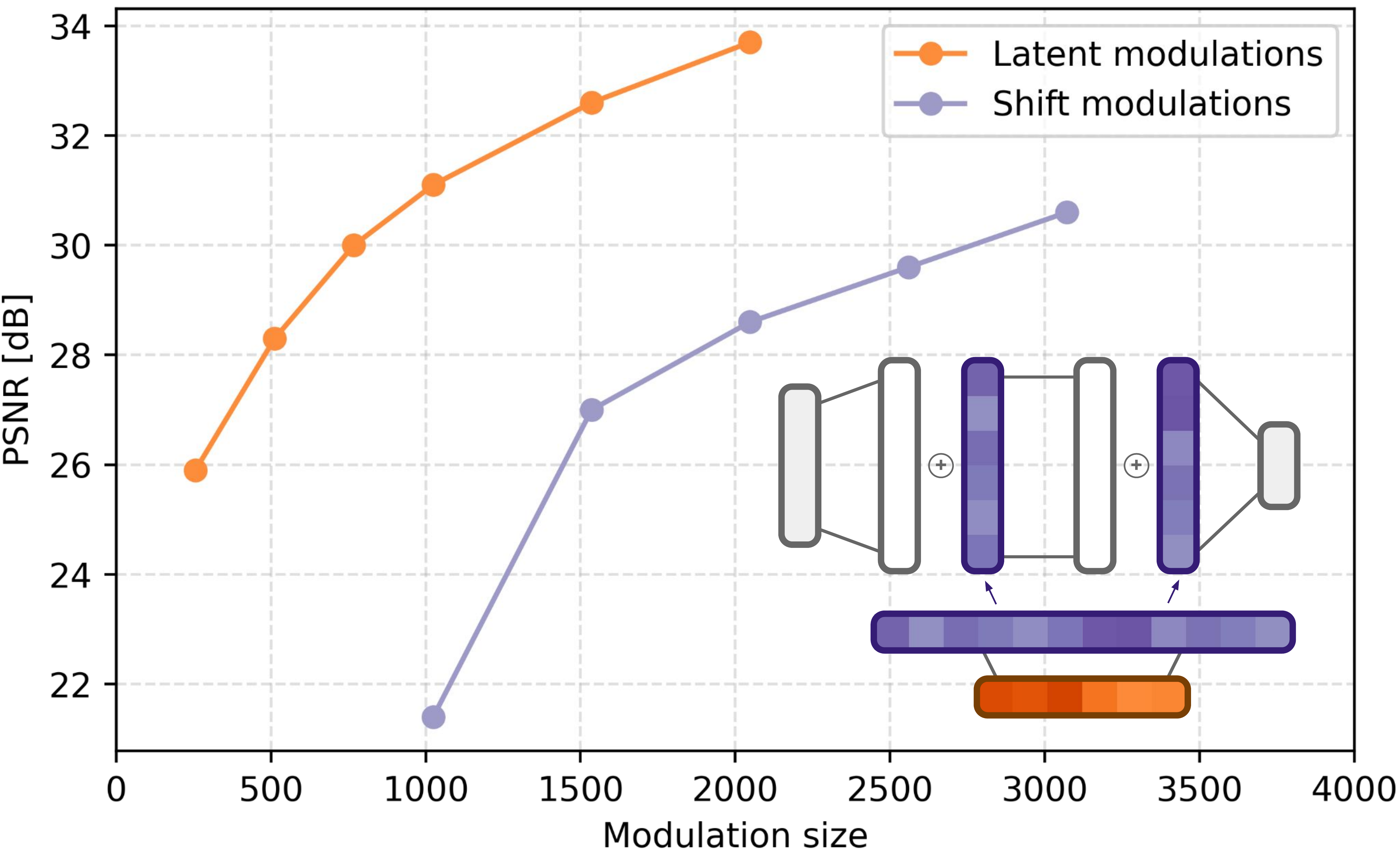}
\end{center}
\vspace{-8pt}
\caption{Reconstruction accuracy (in PSNR) vs modulation dimensionality on CelebA-HQ 64$\times$64. Reconstruction accuracy is computed from the MSE between the image array and its INR evaluated at each pixel location. The model architecture is shown on the bottom right, with purple vectors corresponding to shift modulations and orange vectors to latent modulations.}
\label{fig:rate-distortion}
\vspace{-3mm}
\end{figure}

We can further reduce the number of modulations by using \textit{latent modulations}, a vector that is linearly mapped to the shift modulations that is similar to the generator in \citet{chan2020pi} (\autoref{fig:rate-distortion}). Then instead of storing the shift modulations, we can store this latent vector to represent any given data point (with the parameters used for mapping latents to shift modulations being part of the base network shared across data points). \autoref{fig:rate-distortion} shows that this approach results in the best tradeoff between reconstruction accuracy (\autoref{eq:inr-opt}) and the dimensionality of the modulations. We therefore use this architecture for the remainder of the paper and also use the word \textit{modulations} to refer to latent modulations. We typically use modulation dimensions of 256 and 512 as these result in reconstructions that are visually very close to the original data, while being orders of magnitude smaller than array representations. These modulations are then used as inputs for the downstream deep learning tasks that we describe in \Cref{sec:dl-on-functa}. We also experimented with other architectures for representing functa but found that they performed worse (\textit{cf.}\ \Cref{sec:modulation-architectures}).

\subsection{Meta-learning functa}

Given the representation of each functa point as a modulation applied to a base network, we are now faced with two problems:
1.\ How do we learn the weights of the base network such that they efficiently encode the shared characteristics of data points?
2.\ Fitting INRs can be slow -- e.g.\ fitting a single NeRF scene can take 1 GPU day. If we are to create a dataset of several thousand NeRF scenes (as we do in this paper), fitting each of them individually would be prohibitively expensive. How can we efficiently create large datasets of modulations?

We solve both problems by meta-learning a network initialization such that each functa point can be fitted in a few gradient steps (\textit{cf.}\ Algorithm \autoref{alg:meta-learning}). Indeed, \citet{tancik2021learned} show that applying MAML \cite{finn2017model} to INRs can enable fitting of images and NeRF scenes in only a handful of gradient steps. Similar results were also shown previously on signed distance functions \cite{sitzmann2019metasdf}. However in both works, the initialization of all neural network parameters is meta-learned then fitted to each data point. In our case each functa is stored as a modulation vector, and when creating the functaset we only fit the modulation to each data point with the shared base network fixed. In the inner loop we therefore only update the modulations, whereas in the outer loop we only update the base network weights. 
This then corresponds to an instance of learning a subset of weights with MAML, also known as CAVIA \cite{zintgraf2019fast}.

After meta-learning the base network using the training data, we create a dataset of modulations by running the inner loop on each data point that takes a few gradient steps to fit the modulation. We found that using 3 gradient steps works well for both meta-learning and fitting modulations. This is done for each data point of the training set and test set, to obtain modulation datasets for both training and test. 
An additional benefit of meta-learning the weights and modulations is that it makes the modulation space smooth, as all modulations are a handful of gradient steps away from each other. We hypothesize that this simplifies downstream tasks, yet also faces limitations.
We discuss the limitations of meta-learning and alternatives in \Cref{sec:limitation-future-work}.

\begin{algorithm}[t]
\caption{Meta-learning functa}
\label{alg:meta-learning}
\begin{algorithmic}[1]
\STATE Randomly initialize shared base network $\theta$
\WHILE{not done}
\STATE Sample batch $\mathcal{B}$ of data $\{\{\mathbf{x}_i^{(j)}, \mathbf{f}_i^{(j)}\}_{i \in \mathcal{I}}\}_{j \in \mathcal{B}}$
\STATE Set batch modulations to zero $\phi_j \leftarrow 0$ $\forall j \in \mathcal{B}$
 \FORALL{step $\in \{1, \ldots, N_{inner} \}$ and $j \in \mathcal{B}$}
 \STATE $\phi_j \leftarrow \phi_j - \epsilon \nabla_\phi \mathcal{L}(f_{\theta,\phi}, \{\mathbf{x}^{(j)}_i, \mathbf{f}^{(j)}_i\}_{i \in \mathcal{I}})|_{\phi=\phi_j}$
 \ENDFOR
 \STATE $\theta \leftarrow \theta - \epsilon' \frac{1}{|\mathcal{B}|}\sum_{j \in \mathcal{B}} \nabla_\theta \mathcal{L}(f_{\theta,\phi}, \{\mathbf{x}^{(j)}_i, \mathbf{f}^{(j)}_i\}_{i \in \mathcal{I}})|_{\phi=\phi_j}$
\ENDWHILE
\end{algorithmic}
\end{algorithm}

\section{Deep learning on functa} \label{sec:dl-on-functa}

Given the modulation representation of functa, we train deep learning models directly on the modulations for a selection of downstream tasks. This decoupling of fitting functa and the downstream task is more memory-efficient than joint learning, and modular in the sense that we can use the same functa for different tasks. Also note that modulations are arrays, but are different to the original array data and conventional latent representations in deep learning in that they parameterize a function, hence enjoy the advantages of functa listed in \cref{sec:functa}.

We mainly focus on the task of generative modeling not only because it is challenging, but also because it is a broad task that can be used for inference, which includes applications such as imputation and novel view synthesis as we later describe in this section. We also show results on classification to demonstrate the scope of the framework, covering both generative and discriminative tasks.

\textbf{Generative modelling}. We primarily use normalizing flows \cite{rezende2015variational, durkan2019neural} and diffusion \cite{sohl2015deep, ho2020denoising} for generative modelling, chosen based on ease of optimization. VAEs \cite{kingma2013auto, rezende2014stochastic} and GANs \cite{goodfellow2014generative} are also sensible choices that we do not explore in this paper. We also explored Transformers \cite{vaswani2017attention} as an example of an autoregressive generative model, but found them to underperform (\textit{cf.}\ \autoref{sec:didnt-work}). This may be because it is difficult to impose a meaningful ordering to the modulation dimensions.

Normalizing flows model the data distribution by mapping a simple base distribution (usually a Gaussian) through a sequence of invertible layers parameterized by neural networks. Diffusions model the data by applying a sequence of fixed Gaussian noise (forward process) to the data, and then training a neural network to denoise the noisy version of the data at each step (backward process). We use Neural Spline Flows (NSF) \cite{durkan2019neural} and DDPM \cite{ho2020denoising} with MLP-based architectures as an example of each generative model, providing background and implementation details in \Cref{sec:nsf} and \Cref{sec:ddpm}.

Each flow/diffusion layer preserves the dimensionality of data that it is trained on, so the model size typically scales with input dimensionality. For these models, it is therefore much more efficient to train on modulations compared to the original array representation. In the case of NeRF scenes, there is no obvious way to form an array representation of each scene on which we can train flows or diffusion. In contrast, when representing functa as modulations we can directly learn the distribution of modulations, greatly simplifying generative modeling of NeRF scenes.

\textbf{Inference}. Given a generative model over modulations, we can formulate various applications as an inference problem. The learned distribution over modulations can be interpreted as the prior, and the reconstruction loss of the functa corresponding to the modulation can be interpreted as the likelihood. By optimising a weighted average of this prior and likelihood with respect to the modulations $\phi$, we can perform MAP inference: 
\begin{equation} \label{eq:inference}
    \min_{\phi} - \log p(\phi) + \lambda \sum_{i \in \mathcal{I}} \| f_\phi(\mathbf{x}_i) - \mathbf{f}_i \|^2_2
\end{equation}
This can be used for imputation when the likelihood is computed on a partial observation. For example, we can optimise a modulation to achieve high log probability under the prior as well as fitting one half of a voxel grid. We then query the INR represented by the resulting modulation at all grid points to fill in the other half (\textit{cf.}\ \autoref{fig:chair-imputation}). In terms of \autoref{eq:inference}, $\mathcal{I}$ would then correspond to coordinates of one half of the voxel grid. Similarly if the reconstruction error is computed on a (partial) view of a scene, then MAP inference will allow us to infer the scene by fitting the modulations via \autoref{eq:inference}. The scene can then be rendered at arbitrary viewpoints for novel view synthesis. Note that flows are more suitable as the prior for inference than diffusion, as flow densities can be computed exactly and efficiently whereas diffusion densities are usually intractable or expensive to compute.

\textbf{Classification}.
We also train classifiers directly on modulation datasets. % the voxel modulation datasets. 
With only small MLPs, we can reach high test accuracy in a few thousand iterations (minutes of wall clock time on a single GPU).

%%%%%%%%%%%%%%%%%%%%%%%%%%%%%%%%%%%%%%%%%%%%%%%%%%

\section{Related Work}

\textbf{Generative models of functions}. In early work, \citet{ha2016generating} trained GANs and VAEs on functional representations of MNIST. Later, Neural Processes \citep{garnelo2018conditional, garnelo2018neural,kim2019attentive, gordon2019convolutional} were introduced to model conditional distributions of 1D functions and images as 2D functions. \citet{skorokhodov2021adversarial, anokhin2021image} introduce an adversarial approach for learning distributions of INRs for images, generating high quality images. INRs have also been used to learn distributions of 3D shapes using GANs \cite{chen2019learning,kleineberg2020adversarial}, VAEs \cite{mescheder2019occupancy} and score-based generative models \cite{cai2020learning}. While not formally generative models, auto decoders have also been used to parameterize families of 3D shapes \cite{park2019deepsdf, atzmon2020sald}. A series of recent works have built generative models for NeRF scenes mostly using GANs \cite{schwarz2020graf, niemeyer2020giraffe, chan2020pi, chan2021efficient, devries2021unconstrained} or VAEs \cite{kosiorek2021nerf}.

The most closely related works to ours are GASP \cite{dupont2021generative} and GEM \cite{du2021gem}. GASP learns distributions of INRs using a modality agnostic point-cloud discriminator with a GAN for learning distributions of images, 3D shapes and data on manifolds. However, unlike our approach, GASP's GAN-based training is unstable and not applicable to NeRF scenes as it is unclear how to feed in a scene to the discriminator. GEM learns a manifold of INRs in a modality agnostic manner, by embedding latent vectors (mapped to INRs through a hypernetwork) into a space which is regularized to have various desirable properties. GEM is successfully applied across a range of modalities and a variety of tasks, but the learned embeddings are not used to train generative models or classifiers.

\textbf{Multimodal architectures}. The literature on multimodal processing usually relies on modality-specific feature extractors \cite{kaiser2017one, chen2019uniter, alayrac2020self}. However the recently introduced Perceiver \cite{jaegle2021perceiver, jaegle2021perceiverio} uses a shared achitecture for processing a wide range of data modalities, sharing similarities with our setup. However they are only applied to array representations of data, and it would be an interesting research direction to apply them to functa for downstream tasks.

\textbf{Diffusion and flows in latent space}. There has also been a variety of work on applying diffusion to the latent space of a VAE \cite{vahdat2021score, mittal2021symbolic, wehenkel2021diffusion, sinha2021d2c}. Similarly there have been various works that use flow priors for VAEs \cite{chen2016variational, huang2017learnable, xiao2019generative}. These are in contrast to our work that applies diffusion and flows to functional representations.

\textbf{Deep learning on neural networks.} There have been a few works where neural networks are used as inputs to other neural networks. \citet{unterthiner2020predicting} predict classification accuracies of CNNs directly from their vectorized weights. \citet{schurholt2021self} further apply self-supervised learning to the vectorized weights and use the resulting representations to predict various characteristics of the input classifier. \citet{knyazev2021parameter, jaeckle2021generating, lu2019neural} represent the computational graph of neural networks as a GNN that is used to predict optimal parameters, adversarial examples, or branching strategies for neural network verification. These works differ from ours in that their input neural networks do not represent functions, and their goal is to predict quantities about the neural network rather than to perform generative/discriminative tasks.

%%%%%%%%%%%%%%%%%%%%%%%%%%%%%%%%%%%%%%%%%%%%%%%%%%%%%%%%%%%%%%%%%%%%%

\section{Experiments}

We evaluate our framework on four data modalities: \textit{images}, using the CelebA-HQ 64$\times$64 dataset \cite{karras2018progressive}, \textit{voxels}, using the ShapeNet dataset \cite{chang2015shapenet}, \textit{NeRF scenes}, using the SRN Cars dataset \cite{sitzmann2019scene} and \textit{ data on manifolds} using the ERA5 temperature dataset \cite{hersbach2018era5}. We implement all models in Jax \cite{jax2018github} using Haiku \cite{haiku2020github} and Jaxline \cite{deepmind2020jax} for training. We discuss negative results in \Cref{sec:didnt-work}. Code for the meta-learning experiment for CelebA-HQ 64$\times$64 and SRN Cars, and for creating a modulation dataset for CelebA-HQ 64$\times$64 is opensourced at: \href{https://github.com/deepmind/functa}{\nolinkurl{github.com/deepmind/functa}}.

\subsection{Meta-learning}

\begin{figure}[t]
\hspace{14pt} Init \hspace{24pt} Step 1 \hspace{16pt} Step 2 \hspace{16pt} Step 3 \hspace{16pt} Target
\begin{center}
\includegraphics[width=\columnwidth]{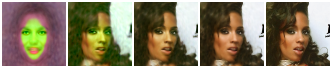}
\includegraphics[width=\columnwidth]{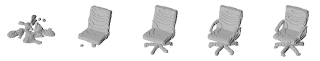}
\includegraphics[width=\columnwidth]{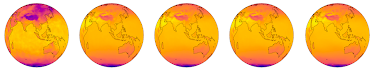}
\includegraphics[width=\columnwidth]{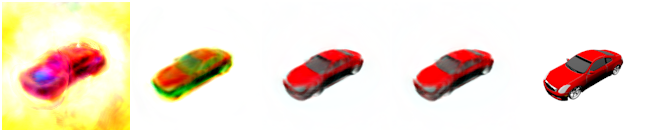}
\end{center}
\vspace{-8pt}
\caption{Visualization of the meta-learned initialization, each gradient step of the inner loop and target data point. GIF showing scenes at different poses: \href{https://github.com/deepmind/functa\#figure-4}{\nolinkurl{github.com/deepmind/functa\#figure-4}}}
\label{fig:meta-learning-visualization}
\vspace{-3mm}
\end{figure}

We first create functasets for each data modality by meta-learning an initialization and then fitting modulations to each data point. As shown in \autoref{fig:meta-learning-visualization}, we are able to fit modulations to a high degree of accuracy in only 3 gradient steps across a range of data modalities. \autoref{tab:mod-rec-acc} shows that we are able to capture signals accurately using only 64-512 modulations, with accuracy usually increasing with modulation size. The one exception is SRN Cars, where reconstruction quality is similar across all modulation sizes. We suspect this is due to our very basic rendering scheme (\textit{cf.}\ \Cref{sec:nerf-scenes}).
Qualitative examples of reconstructions for each data modality are shown in \autoref{fig:orig-vs-rec} of \Cref{sec:further-experiments}.

\newcommand{\shaderowshapenetchairs}{\rowcolor{LimeGreen!20}}
\newcommand{\shaderowshapenetten}{\rowcolor{ForestGreen!20}}
\newcommand{\shaderowceleba}{\rowcolor{BurntOrange!20}}
\newcommand{\shaderowsrncars}{\rowcolor{Lavender!30}}
\newcommand{\shaderowera}{\rowcolor{TealBlue!20}}
\begin{table}[t]
    \setlength\tabcolsep{2pt}
    \centering
    \small
    \begin{tabular}{l|l|ccccc}
       \toprule[1pt]
        \multirow{2}{*}{Dataset, array size} & \multirow{2}{*}{Split} & \multicolumn{5}{c}{Modulation dimensionality} \\
         & & 64 & 128 & 256 & 512 & 1024 \\ \midrule
        \shaderowshapenetchairs & Train & 99.43 & 99.49 & 99.49 & 99.51 & 99.53  \\ 
        \shaderowshapenetchairs \multirow{-2}{*}{ShapeNet Chairs, $64^3$} & Test & 99.11 & 99.28 & 99.38 & 99.46 & 99.51 \\ 
        \shaderowshapenetten & Train & 99.36 & 99.44 & 99.47 & 99.52 & 99.56  \\ 
        \shaderowshapenetten \multirow{-2}{*}{ShapeNet 10 Classes, $64^3$} & Test & 99.30 & 99.40 & 99.44 & 99.50 & 99.55 \\ 
        \shaderowceleba  & Train & 22.2 & 24.2 & 26.6 & 29.7 & 32.4 \\ 
        \shaderowceleba \multirow{-2}{*}{CelebA-HQ, 64$\times$64} & Test & 21.6 & 23.5 & 25.6 & 28.0 & 30.7 \\
        \shaderowsrncars & Train & 24.3 & 24.2 & 24.6 & 24.6 & 24.4 \\ 
        \shaderowsrncars \multirow{-2}{*}{SRN Cars, 128$\times$128} & Test & 22.4 & 23.0 & 23.1 & 23.2 & 23.1\\ 
        \shaderowera & Train & 43.2 & 43.7 & 43.8 & 44.0 & 44.1  \\ 
        \shaderowera \multirow{-2}{*}{ERA5, 181$\times$360} & Test & 43.2 & 43.6 & 43.8 & 43.9 & 44.0\\ 
        \bottomrule[1pt]
    \end{tabular}
    \vspace{-8pt}
    \caption{Mean reconstruction of modulations across each dataset vs modulation size. Metric is voxel accuracy (\%) for ShapeNet and PSNR (dB) for the rest. See \Cref{sec:meta-learning-details} for details on metric.}
    \label{tab:mod-rec-acc}
    \vspace{-5mm}
\end{table}

\subsection{Images} \label{sec:images}

\begin{figure}[t]
\begin{center}
\raisebox{33pt}{\rotatebox[]{90}{GASP}}\hspace{5pt}
\includegraphics[width=0.93\columnwidth]{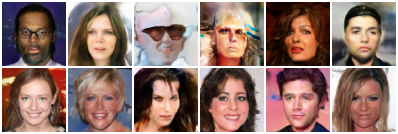}

\raisebox{33pt}{\rotatebox[]{90}{Ours}}\hspace{5pt}
\includegraphics[width=0.93\columnwidth]{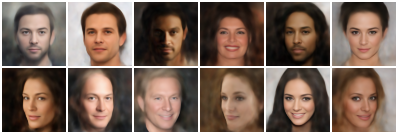}
\end{center}
\vspace{-15pt}
\caption{Uncurated samples from GASP and DDPM (diffusion) trained on 256-dim CelebA-HQ 64$\times$64 modulations.}
\label{fig:celeba-diffusion-samples-256}
\vspace{-5mm}
\end{figure}

We train DDPM \cite{ho2020denoising} on the CelebA-HQ 64$\times$64 functaset with 256 dimensional modulations. As can be seen in \autoref{fig:celeba-diffusion-samples-256}, the diffusion model is able to produce realistic and convincing samples even though it is trained directly on modulations. Compared to GASP, the samples produced by our model are more coherent, although slightly blurrier. To verify that our model has not memorized the training dataset, we show nearest neighbor samples in \autoref{fig:celeba-diffusion-samples-256-knn} in the appendix. Our model achieves an FID score \cite{heusel2017gans} of 40.4, but we found that this was not well correlated with perceptual quality. As noted by \citet{du2021gem}, this is likely due to FID's property of over-penalizing blurriness (see \Cref{sec:fid} for discussion).

\subsection{Voxels}

\begin{figure}[t]
\begin{center}
\raisebox{75pt}{\rotatebox[]{90}{\hspace{4pt} Ours \hspace{11pt}
GASP \hspace{15pt} SD \hspace{20pt} ON \hspace{2pt}}}\hspace{5pt}\includegraphics[width=0.93\columnwidth]{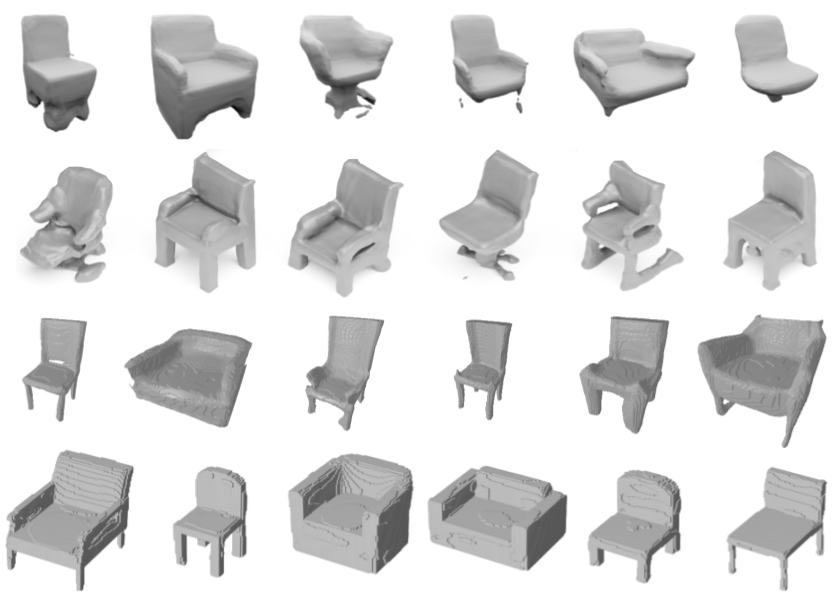}
\end{center}
\vspace{-12pt}
\caption{Unconditional samples from our model and three baselines: Occupancy Networks trained as a VAE (ON) from \citet{mescheder2019occupancy}, a GAN trained on signed distance functions with a set discriminator (SD) from \citet{kleineberg2020adversarial} and GASP from \citet{dupont2021generative}.}
\label{fig:chair-baselines}
\end{figure}

\begin{figure}[t]
\raisebox{25pt}{\rotatebox[]{90}{Observed}}\hspace{5pt}
\includegraphics[width=0.22\columnwidth]{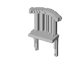}
\includegraphics[width=0.22\columnwidth]{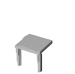}
\includegraphics[width=0.22\columnwidth]{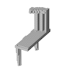}
\includegraphics[width=0.22\columnwidth]{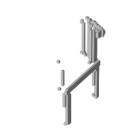}

\raisebox{25pt}{\rotatebox[]{90}{Inferred}}\hspace{5pt}
\includegraphics[width=0.22\columnwidth]{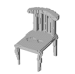}
\includegraphics[width=0.22\columnwidth]{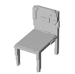}
\includegraphics[width=0.22\columnwidth]{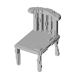}
\includegraphics[width=0.22\columnwidth]{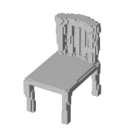}

\vspace{-8pt}
\caption{Imputation from partial test observations using the learned distribution over functa as a prior. GIF showing course of optimization: \href{https://github.com/deepmind/functa\#figure-7}{\nolinkurl{github.com/deepmind/functa\#figure-7}}. Additional chairs: \autoref{fig:chair-imputation-appendix}}
\label{fig:chair-imputation}
\vspace{-3mm}
\end{figure}

\begin{figure}[t]
\begin{center}
\includegraphics[width=\columnwidth]{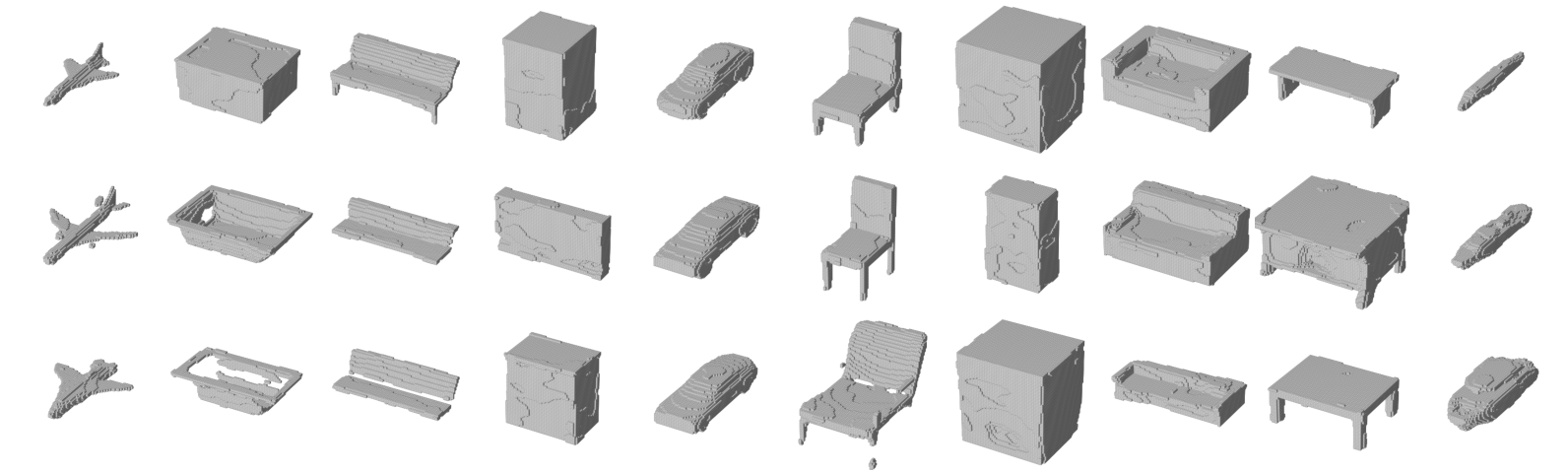}
\end{center}
\vspace{-12pt}
\caption{Uncurated samples from a class-conditional flow trained on 256-dim modulations of ShapeNet 10 Classes $64^3$.}
\label{fig:shapenet-conditional-flow-samples-256}
\end{figure}

We use two datasets derived from the ShapeNet database \cite{chang2015shapenet}, $64^3$ voxels from the chairs category and $64^3$ voxels from 10 classes, using data augmentation to ensure each class contains a large number of samples (\textit{cf.}\ \Cref{sec:datasets}). We train NSF \cite{durkan2019neural} on a 256 dimensional functaset of the chairs category, with unconditional samples shown in \autoref{fig:chair-baselines} along with comparisons to other baseline generative models using 3D functional representations. As can be seen, our model generates coherent and realistic samples, whereas the baselines tend to be less consistent. Further, our approach stably and consistently produces good results whereas GASP is unstable to train for 3D voxels \cite{dupont2021generative}. Our model can also be used to produce semantically meaningful latent interpolations as shown in \autoref{fig:chair-latent-interpolation} in \Cref{sec:further-experiments}. \autoref{fig:shapenet-conditional-flow-samples-256} shows samples from a class-conditional NSF trained on 10 classes from ShapeNet, where our model produces realistic and consistent samples for each class. Finally, we show imputation results in \autoref{fig:chair-imputation}, where our model is able to infer the shape of the chair from various partial observations, including a simulated lidar scan (last column).

\subsection{NeRF scenes} \label{sec:nerf-scenes}

\begin{figure}[t]
\begin{center}
\raisebox{38pt}{\rotatebox[]{90}{$\pi$-GAN - \small{\textit{FID: 36.7}}}}\hspace{5pt}
\includegraphics[width=0.93\columnwidth]{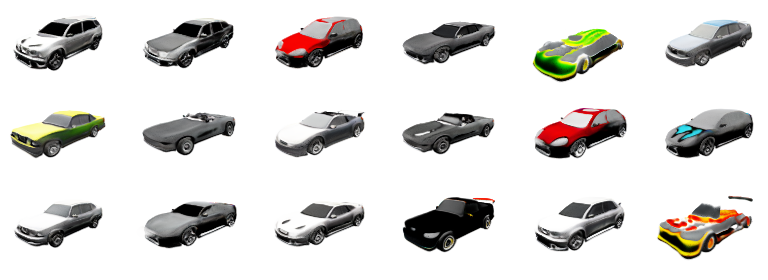}

\vspace{8pt}

\raisebox{40pt}{\rotatebox[]{90}{Ours - \small{\textit{FID: 80.3}}}}\hspace{5pt}
\includegraphics[width=0.93\columnwidth]{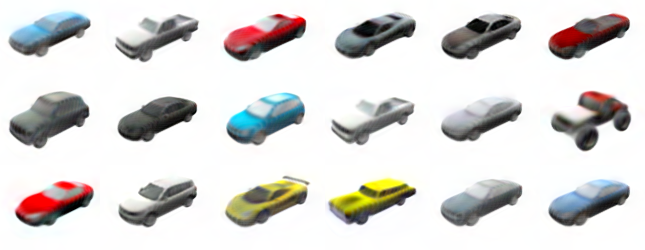}
\end{center}
\vspace{-15pt}
\caption{Uncurated samples from $\pi$-GAN and DDPM trained on 64-dim modulations of SRN Cars. GIF showing different poses: \href{https://github.com/deepmind/functa\#figure-9}{\nolinkurl{github.com/deepmind/functa\#figure-9}}}
\label{fig:car-ddpm-samples-64}
\vspace{-3mm}
\end{figure}

\begin{figure}[t]
\begin{center}
\includegraphics[width=\columnwidth]{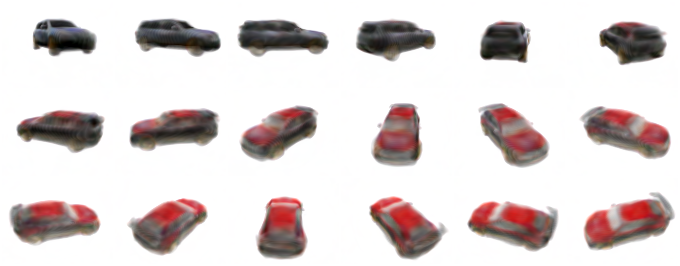}
\end{center}
\vspace{-8pt}
\caption{Latent interpolation between two car scenes with moving pose. GIF: \href{https://github.com/deepmind/functa\#figure-10}{\nolinkurl{github.com/deepmind/functa\#figure-10}}}
\label{fig:car-flow-latent-interpolation-512}
%\vspace{-5mm}
\end{figure}

\begin{figure}[t]
\includegraphics[width=1.0\columnwidth]{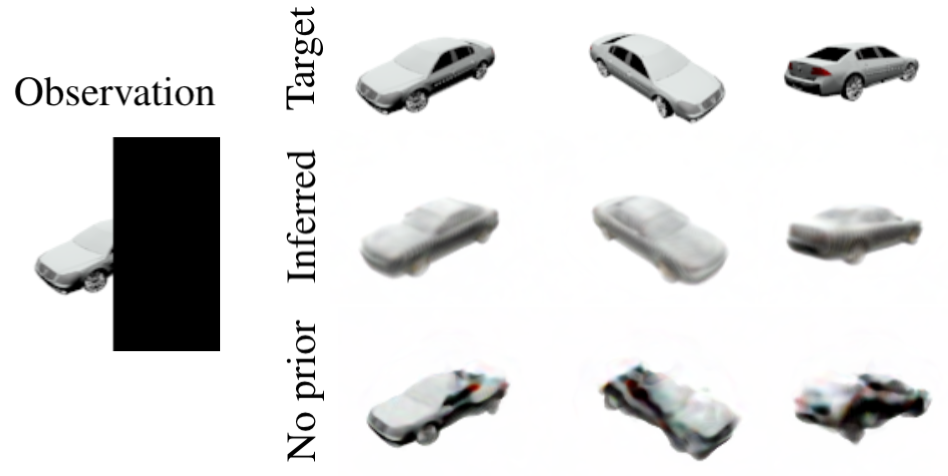}
\vspace{-12pt}
\caption{Novel view synthesis from occluded test scene. Without the prior, the model is not able to correctly infer the shape of the car. GIF: \href{https://github.com/deepmind/functa\#figure-11}{\nolinkurl{github.com/deepmind/functa\#figure-11}}. Additional scenes: \autoref{fig:novel-view-synthesis-appendix}}
\label{fig:novel-view-synthesis}
\vspace{-3mm}
\end{figure}

We evaluate our framework on NeRF scenes by using the SRN Cars dataset \cite{sitzmann2019scene}, containing 2458 scenes each with 50 posed images of size 128$\times$128. NeRF represents a scene via an INR mapping 3D coordinates to density and RGB values (\textit{cf.}\ \autoref{fig:figure1}). Views of the scene are then generated via volume rendering, and modulations are fitted by minimizing reconstruction loss on the available set of posed images on training scenes (see \Cref{sec:experimental-details} for background and details). While NeRF uses several tricks to achieve good performance, we use a minimal bare bones model for our experiments (\textit{cf.}\ \Cref{sec:nerf-details}), as we are more interested in testing the generative capabilities of our model rather than the visual quality of NeRF. As a baseline, we compare against $\pi$-GAN \cite{chan2020pi} using model weights kindly provided to us by the authors.

We train both DDPM and NSF on NeRF modulations, achieving similar sample quality. As shown in \autoref{fig:car-ddpm-samples-64}, DDPM is able to generate plausible NeRF scenes of cars, giving an FID score of 80.3 (\textit{cf.}\ \Cref{sec:datasets} for details on FID computation) compared to 36.7 for $\pi$-GAN. While the samples produced by $\pi$-GAN are generally sharper than ours (as reflected by the FID scores), we believe this can be mitigated by using more sophisticated rendering (including e.g. the hierarchical sampling used by $\pi$-GAN). Further, we note that training $\pi$-GAN requires backpropagating through the volume rendering step, making it very memory intensive. In contrast, we train our model directly on modulations which is much less expensive.

As shown in \autoref{fig:car-flow-latent-interpolation-512}, NSF gives smooth interpolations both in terms of shape and texture. We also perform novel view synthesis experiments in \autoref{fig:novel-view-synthesis}. Given a single occluded view of a test scene, the NSF prior is used to infer a scene that is consistent with the occluded view. Without the prior term in the loss, i.e. when only fitting the modulation to the occluded view, the resulting scene is not realistic, highlighting the importance of using the prior for inference.

% TODO: As the plenoxels work does use arrays + interpolation, we should add a note here along the lines of there is no "compact array based rep" for NeRF scenes.
We highlight that for scenes there is no obvious way to form an array based representation. GAN based models of NeRF scenes \cite{schwarz2020graf, niemeyer2020giraffe} therefore require complex generators and discriminators that backpropagate through the expensive volume rendering operation. Training a generative model in our case is, in contrast, very simple: we simply train a flow or diffusion directly on the modulations. As the generative model itself is independent of the volume rendering step, we believe this holds promise for scaling generative scene models to larger scales than currently possible. In particular, we used a very simple rendering model (no hierarchical sampling, single shared network for density and RGB, no view dependence) limiting the perceptual quality of the renderings. Using more sophisticated rendering techniques \cite{mildenhall2020nerf, barron2021mip} is likely to significantly improve reconstruction and sample quality.

\subsection{Manifold data}

\begin{figure}[t]
\begin{center}
\raisebox{25pt}{\rotatebox[]{90}{GASP}}\hspace{5pt}
\includegraphics[width=0.93\columnwidth]{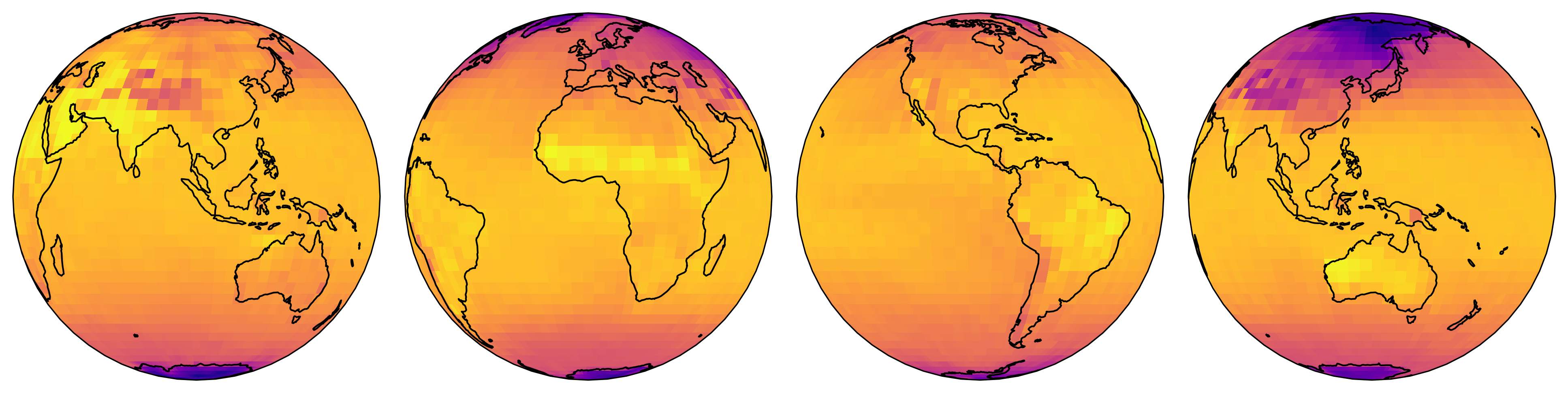}
\raisebox{25pt}{\rotatebox[]{90}{Ours}}\hspace{5pt}
\includegraphics[width=0.93\columnwidth]{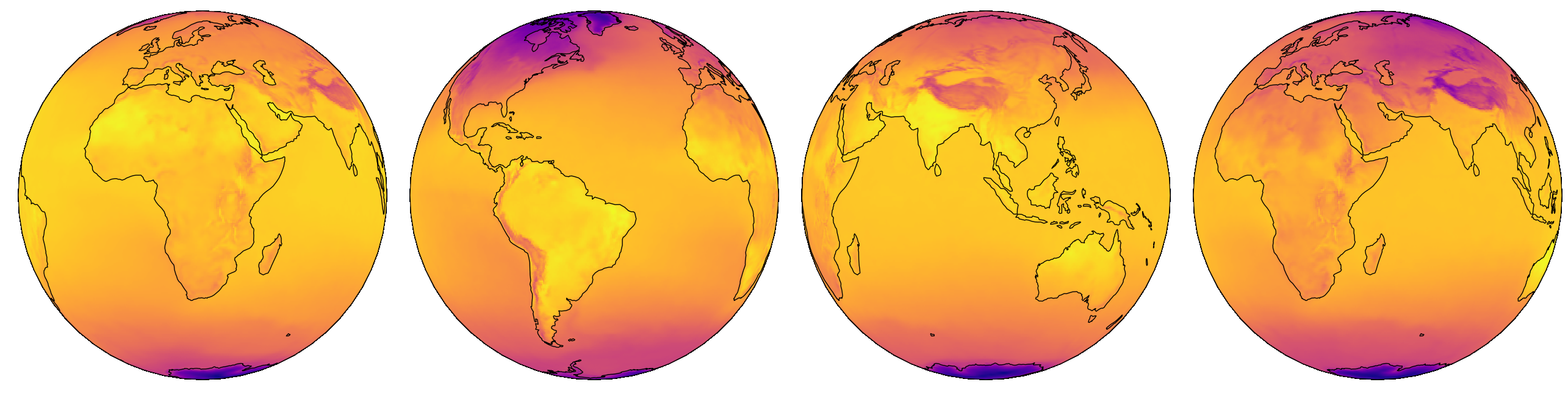}
\vspace{-20pt}
\end{center}
\caption{Uncurated samples from GASP and a flow trained on 256-dim modulations of ERA5. GIF: \href{https://github.com/deepmind/functa\#figure-12}{\nolinkurl{github.com/deepmind/functa\#figure-12}}}
\label{fig:era5-flow-samples-256}
\vspace{-5mm}
\end{figure}

To demonstrate the flexibility of our approach, we also train NSF on 256-dimensional modulations of ERA5 temperature data \cite{hersbach2018era5}. The dataset contains 10k grids of temperature measurements at equally spaced latitudes $\lambda$ and longitudes $\varphi$. We convert them to Cartesian coordinates $\mathbf{x} = (\cos \lambda \cos \varphi, \cos \lambda \sin \varphi, \sin \lambda)$ for the functa inputs as in \citet{dupont2021generative}. As our method is more stable and scalable than GASP, we are able to train on 181$\times$360 grids as opposed to the 46$\times$90 grids used by GASP, yielding higher resolution samples (\autoref{fig:era5-flow-samples-256}).

\subsection{Classification} \label{sec:classification}

\begin{table}[h!]
\begin{center}
\begin{small}
\begin{sc}
\begin{tabular}{lcc}
\toprule
Classifier & Test accuracy & $n_{\text{params}}$ \\
\midrule
MLP on Functa    & $93.6 \pm 0.1\%$ & $83$k \\
3D CNN    & $93.3 \pm 0.3\%$ & $550$k \\
\bottomrule
\end{tabular}
\end{sc}
\end{small}
\end{center}
\caption{Classification accuracies and parameter count for MLP on functa vs 3D CNN on array data for ShapeNet 10 Classes, $64^3$.}
\label{tab:classification}
\vspace{-3mm}
\end{table}

Finally, we evaluate our model for classification using the ShapeNet 10 classes dataset. Given the functaset, the task is to predict the class of the object from the modulation. As a baseline we train a 3D CNN classifier on $64^3$ voxels with an architecture based on \citet{maturana2015voxnet}. As shown in \autoref{tab:classification}, our 4 hidden layer MLP of width 128 performs similarly (if not better) than the 8-layer ($6 \times$ 3DConv + $2 \times$ linear layers) 3D CNN baseline at convergence. Our small MLP can be trained in $< 10$ minutes on a single GPU with batch size 1024. In contrast, the 3D CNN baseline is memory intensive and thus we run on 8 devices each with batch size 128 (to match batch size) for $> 1$ hour, highlighting the scalability of performing deep learning tasks on functa. See \Cref{sec:classification-details} for experimental details and \Cref{sec:further-experiments} for further results and training curves.

\section{Conclusion, limitations and future work} \label{sec:limitation-future-work}

\textbf{Conclusion}. Motivated by the various compelling properties of functional representations, we propose to view INRs as data points, or functa, and treat these as first class citizens for machine learning tasks. We introduced a method for creating such datasets of functa at scale and showed that it is possible to learn generative models, perform data imputation, novel view synthesis and classification across a wide range of data modalities within this framework.

\textbf{Limitations and future work}. Using a shared framework for different data modalities implies that we cannot employ modality specific inductive biases when designing models trained on functa. Indeed, storing functa as modulations removes spatial structure from the data, forcing us to use general MLP architectures. We see promise in spatial functional representations (e.g.\ per-patch modulations in \citet{mehta2021modulated}) on which for example convolutions can be applied, exploiting their locality and translation invariance.

Further, we rely on a variant of MAML to create our functaset, and therefore inherit the limitations of MAML, including a large memory footprint and occasionally unstable training due to the double loop optimization. In addition, MAML constrains the modulations to lie within a few gradient steps of the meta-learned initialization which could limit reconstruction accuracy for more complex datasets. As MAML is a bottleneck in terms of memory and possibly representational power, it would be interesting to explore alternatives to meta-learning for creating functasets, for example using the autodecoder framework \cite{park2019deepsdf}.

When creating the functaset, we must make multiple choices such as the architecture of the base network and the modulation dimensionality. Our metric for this choice was compressibility, i.e.\ using the smallest number of modulations for a given PSNR. However, this may be suboptimal for the downstream task, and an alternative is to use a task-specific metric and/or jointly learn the INR and the downstream model end-to-end. However this may harm the reconstruction quality of the INR, be memory expensive, and would require retraining separate INRs for separate tasks.

The field of INRs is progressing rapidly and we will likely be able to take advantage of this progress, including hybrid representations \cite{martel2021acorn}, better activation functions \cite{ramasinghe2021beyond} and reduced memory consumption \cite{huang2021textrm}. There has also been a plethora of work on improving NeRF \cite{barron2021mip, reiser2021kilonerf, yu2021plenoctrees, piala2021terminerf}, which should be directly applicable to our framework. Further, recent works have shown promise for storing compressed datasets as functions \cite{dupont2021coin, chen2021nerv, strumpler2021implicit, zhang2021implicit}. Using our framework, it may therefore become possible to train deep learning models directly on these compressed datasets, which is challenging for traditional compressed formats such as JPEG (although image-specific exceptions such as \citet{nash2021generating} exist). In addition, learning distributions of functa is likely to improve entropy coding and hence compression for these frameworks \cite{balle2016end}.

\section*{Acknowledgments}

We thank: Hrushikesh Loya, Milad Alizadeh and Adam Golinski for helpful discussions around meta-learning of modulations; Thu Nguyen-Phuoc for helpful discussions around NeRF; Olivia Wiles for providing DDPM implementation and FID score computation; Conor Durkan and Sander Dieleman for helpful discussions around diffusion; Andy Brock for helpful discussion around implicit neural representations; Andriy Mnih and Yee Whye Teh for general feedback; Eric Ryan Chan for sharing the trained $\pi$-GAN model weights for the NeRF baseline. Emilien gratefully acknowledges his PhD funding from Google DeepMind.

\raggedright
\bibliography{main}
\bibliographystyle{icml2022}

\newpage
\appendix
\onecolumn
\section{Experimental details}\label{sec:experimental-details}

\subsection{Datasets} \label{sec:datasets}
\textbf{CelebA-HQ 64 $\times$ 64}. We use a train/test split of 27,000/3,000. The pixel coordinates $\mathbf{x}_i$ (inputs to SIREN) are normalized to lie in $[0,1]^2$ and pixel intensities $\mathbf{f}_i$ are also normalized to lie in $[0,1]$. More precisely the $\mathbf{x}_i$ are set to be the coordinates of the pixel centers when the image corners are at $\{(0,0),(1,0),(0,1),(1,1)\}$, the vertices of the unit square.

\textbf{ShapeNet}. We take the original $128^3$ voxel dataset and downscale to $64^3$ using \verb!scipy.ndimage.zoom! with threshold=0.05. We found this value of the threshold to be a suitable tradeoff between preserving the structure of the original shapes and having a smooth shape. We also found it important to augment the dataset for preventing overfitting for downstream generative modelling. We apply a 50-fold augmentation by independently rescaling the shape in each of the 3 axes ($x,y,z$) by a randomly sampled scale in $U[0.75, 1.25]$. The resulting Chairs dataset has a train/test split of 304,850/33,900 and the 10 class dataset has a train/test split of 1,516,750/168,850. The voxel grid coordinates $\mathbf{x}_i$ lie in $[0,1]^3$, and the voxel occupancies $\mathbf{f}_i$ are binary, one of $\{0,1\}$. Similarly to images the $\mathbf{x}_i$ are set to be the coordinates of the voxel centers when the full $64^3$ voxel's corners are at the vertices of the unit cube.

\textbf{ERA5 temperature}. This dataset comes with several decades (1979-2020) worth of temperature observations of grids at equally spaced latitudes $\lambda$ and longitudes $\varphi$ across the globe, which we downsample to a $181 \times 360$ grid (from the original $721\times1440$ grid). Note that we treat different time steps as different data points, as modelling the data as a function of time and latitude/longitude would reduce the dataset to a single data point. The $\mathbf{x}_i=(\cos \lambda \cos \varphi, \cos \lambda \sin \varphi, \sin \lambda)$ are 3D Cartesian coordinates obtained from latitudes $\lambda$ that are equally spaced between $\pi/2$ and -$\pi/2$ and longitudes $\varphi$ that are equally spaced between 0 and $\frac{2\pi(n-1)}{n}$ where $n$ is the number of distinct values of longitude (360). We use a train/test split of 9676/2420.

\textbf{SRN Cars}. This dataset has a train/test split of 2458/703 car scenes where each train scene has 50 views (randomly sampled from sphere centered at car) and test scene has 251 views (uniformly distributed on upper hemisphere centered at car). Each view consists of: 
\begin{enumerate}
    \item RGB image of size 128$\times$128$\times$3
    \item Pose information of size 4$\times$4 that contains a 3$\times$3 orthogonal matrix mapping the camera's frame  of reference to world coordinates and 3$ \times$1 coordinates of the camera's position in the world
    \item Scalar focal length of camera
\end{enumerate}
The pose information and focal value are used to calculate the origin and direction of the $128 \times 128$ rays from the camera position to each pixel of the view. Then \verb!num_points_per_ray! points are sampled along each ray at regular intervals, starting from a distance \verb!near! to \verb!far! from the camera. This dataset uses values (\verb!near!, \verb!far!)=$(1.25, 2.75)$. The 3D coordinates of these sampled points on the ray are the $\mathbf{x}_i$ that are fed into a SIREN to produce the corresponding RGB and density value corresponding to that point in 3D space. These values along the ray are combined via volumetric rendering (see details in \Cref{sec:nerf-details}) to compute the prediction of the pixel intensity corresponding to that ray, and the SIREN's parameters are optimized to minimise the reconstruction error at all rays for \verb!num_points! subsampled pixel locations and \verb!num_views! subsampled views (subsampling is necessary to fit the optimization in device memory). These are hyperparameters that are sweeped over for meta-learning (see details below). Regarding FID computation, note that FID is typically calculated with respect to the train set. However for SRN Cars, the train set only contains random views, half of which are looking into the bottom of the car. Thus it is unsuitable to use the train set for sample quality quantification via FID, where we are much more interested in views from the upper hemisphere. Hence we calculate FID score by computing InceptionV3 \citep{szegedy2016rethinking} features on all views of the test set, and comparing against the same views for $n$ sampled scenes where $n$ is the number of test scenes.

\subsection{SIREN and modulations} \label{sec:siren-modulation}
Each layer of SIREN \cite{sitzmann2020implicit} is parameterised as follows:
\begin{equation} \label{eq:siren}
\mathbf{x} \mapsto \sin(\omega_0(\mathbf{W}\mathbf{x}+\mathbf{b}))
\end{equation}
where $\mathbf{W}, \mathbf{b}$ are trainable parameters and $\omega_0$ is a fixed hyperparameter. We fix $\omega_0=30$ for all experiments (except for SRN Cars where we use $\omega_0=5$), as we observed that performance was similar for $\omega_0=\{30,50,70\}$. Following the original SIREN implementation, the weights $\mathbf{W}$ are randomly initialized from $U[-\frac{1}{n_{in}},\frac{1}{n_{in}}]$ for the first layer and $U[-\frac{1}{\omega_0}\sqrt{\frac{6}{n_{in}}}, \frac{1}{\omega_0}\sqrt{\frac{6}{n_{in}}}]$ for subsequent layers, where $n_{in}$ is the input dimensionality of the layer. Note that initialization is different than usual to account for the $\omega_0$ multiplicative term in each SIREN layer. Biases $\mathbf{b}$ are initialized to zero as usual.

ModulatedSIREN is a variant of SIREN that contains a shift modulation $\mathbf{s}$. In our context the SIREN is treated as the base network shared across data points and the shift modulations model the variation across the dataset. For an $n$-layer ModulatedSIREN with weights and biases $\{\mathbf{W}^{(i)}, \mathbf{b}^{(i)}\}_{i=1:n}$, $\mathbf{s}=[\mathbf{s}^{(1)}, \ldots, \mathbf{s}^{(n)}]$ corresponding to the shift modulation $\mathbf{s}$, the $i$th layer is parameterised as:
\begin{equation} \label{eq:siren-modulated}
\mathbf{x} \mapsto \sin(\omega_0(\mathbf{W}^{(i)} \mathbf{x}+\mathbf{b}^{(i)} + \mathbf{s}^{(i)}))
\end{equation}

LatentModulatedSIREN is the variant that we use throughout the paper, which uses a latent modulation vector $\phi$ that is linearly mapped to the shift modulation. So each layer of LatentModulatedSIREN is the same as ModulatedSIREN (\autoref{eq:siren-modulated}), but the shift modulation $\textbf{s}$ is parameterised by $\textbf{s}=\mathbf{W}'\phi + \mathbf{b}'$ for learnable weights $\mathbf{W}'$ and biases $\mathbf{b}'$, both initialized by the Haiku default.

For all SIREN variants, the final layer is simply a linear layer (with no $\omega_0$ scaling nor sine non-linearity), with 0.5 added to the output since the targets are preprocessed to lie in $[0,1]$. For both ModulatedSIREN and LatentModulatedSIREN, the modulations are always initialized to 0 and fitted by a few gradient descent steps of the inner loop, instead of being randomly initialized.

For a given size for the latent modulation for each dataset, we sweep over the LatentModulatedSIREN depth from $\{10, 15, 20\}$ and width from $\{256, 384, 512\}$ and choose the architecture with the best test PSNR/voxel accuracy after meta-learning on the training set.

\subsection{Meta-learning Functa} \label{sec:meta-learning-details}

\begin{table}[h!]
    \centering
    \small
    \begin{sc}
    \begin{tabular}{l|ccccc}
        \toprule[1pt]
        Dataset & batch size per device & num devices & num iterations \\ \midrule
        ShapeNet Chairs  & 1 & 8 & 1e6 \\
        ShapeNet 10 Classes  & 1 & 8 & 1e6 \\
        CelebA-HQ & 16 & 16 & 5e5 \\
        SRN Cars & 1 & 8 & 5e5 \\
        ERA5 temperature  & 2 & 2 & 1e5 \\
        \bottomrule[1pt]
    \end{tabular}
    \end{sc}
    \caption{Selected hyperparameter values for meta-learning functa for each dataset.}
    \label{tab:meta-learning-hyperparams}
\end{table}

For the outer loop, we use Adam \cite{kingma2014adam} with a fixed base learning rate of 3e-6. For the inner loop, we use SGD with learning rate 1e-2. For each modality we use the maximum batch size per device that fits in memory for the biggest LatentModulatedSIREN model we sweep over. In \autoref{tab:meta-learning-hyperparams} we provide hyperparameter values for each dataset.

For SRN Cars, we provide further details about subsampling views and pixels in \Cref{sec:nerf-details}.

Tips for meta-learning:
\begin{itemize}
    \item Training can be unstable for larger SIRENs, in which case it helps to reduce the outer loop learning rate - lowering inner loop learning rate had little effect.
    \item Raising batch size always helps.
    \item Narrow and deep SIREN architectures work better than wide and shallow ones.
    \item Training for many iterations helps.
\end{itemize}

Note that the PSNR values across whole datasets in \autoref{tab:mod-rec-acc}  are computed by first taking the average MSE across whole dataset then converting to PSNR by the formula: $\text{PSNR}= -10 \log_{10}(\text{MSE})$. The voxel accuracy is simply the ratio of correctly reconstructed voxels (after rounding to \{0,1\}) to all voxels, averaged across the dataset.

% We are now interested in fitting the data as well as possible using the lowest number of modulations, as this would imply a lower dimensionality for our functa, leading to compute/memory efficiency for downstream DL task. It would therefore be advantageous to find a better optimization procedure in the inner loop so that the modulations can be fitted more effectively.
As an additional detail, we apply meta-SGD \cite{li2017meta} for our meta-learning i.e. instead of using a fixed learning rate in the inner loop, we learn a learning rate for each parameter (i.e. each modulation). Note that meta-SGD only uses one step to fit their model, whereas we use several steps. Other approaches have also used learnable learning rates for multiple steps, e.g. MAML++ \cite{antoniou2018train} uses per-layer (as opposed to per-parameter) and per-step learning rates. The learning rates are initialised to $U[0.005, 0.1]$ and clipped at $(0,1)$. We found that meta-SGD noticeably improves performance for shift modulations, although didn't make much difference for latent modulations.

\subsection{Neural Spline Flows (NSF)} \label{sec:nsf}
\textbf{Background}. The Neural Spline Flow (NSF) \cite{durkan2019neural} is an example of a normalizing flow \cite{rezende2015variational} that models data $\mathbf{x}$ as the output of a differentiable and invertible transformation $f_\theta$ of Gaussian noise $\mathbf{z}$: 
\begin{equation}
\mathbf{x} = f_\theta(\mathbf{z}), \hspace{3mm} \mathbf{z} \sim p(\mathbf{z})
\end{equation}
where $f_\theta$ is parameterized with a neural network. Note that in our context, the data $\mathbf{x}$ are modulations. The inverse transformation $f_\theta^{-1}$ (that takes in the modulation data) is parameterized by a neural network whose parameters are optimized by maximising the log probability of the data under the flow, that is computed by the change of variable formula:
\begin{equation}
    \log p(\mathbf{x}) = \log p(f_\theta^{-1}(\mathbf{x})) + \log \left\lvert \frac{\partial f_\theta^{-1}}{\partial \mathbf{x}} \right\rvert.
\end{equation}
The invertible transformation $f_\theta$ consists of multiple flow layers (each with a distinct set of parameters) where each flow layer is the composition of a coupling transform and a PLU invertible linear transform.

A \textit{coupling transform} maps an input $\mathbf{x} \in \mathbb{R}^{2d}$ to output $\mathbf{y} \in \mathbb{R}^{2d}$ as follows:
\begin{equation}
    [\mathbf{y}_{1:d}, \mathbf{y}_{d+1:2d}] = [\mathbf{x}_{1:d}, b(\mathbf{x}_{d+1:2d} ; g_{\theta}(\mathbf{x}_{1:d}))]
\end{equation}
where $b:\mathbb{R}^d \rightarrow \mathbb{R}^d$ is an elementwise bijection based on neural splines, whose parameters are given by a conditioner $g_{\theta}(\mathbf{x}_{1:d})$. It is easy to check that for any invertible $b$, the coupling transform is also invertible. This conditioner $g_{\theta}: \mathbb{R}^d \rightarrow \mathbb{R}^{pd}$ is an MLP that outputs the $p$ parameters per input dimension for the neural spline bijector.

\begin{figure}[t]
\begin{center}
\includegraphics[width=0.6\columnwidth]{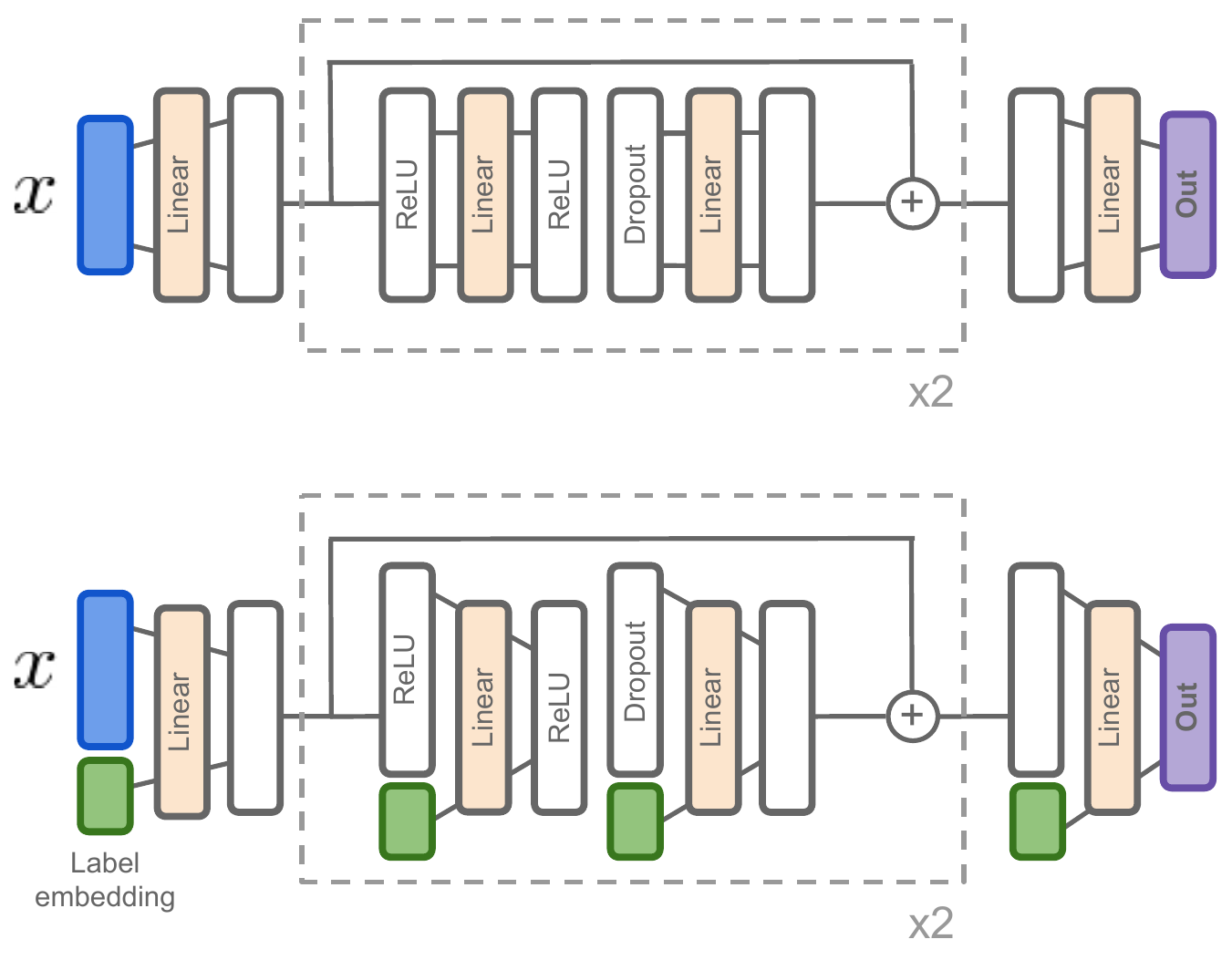}
\end{center}
\caption{Architecture for neural spline flow conditioner $g_\theta$ for the unconditional flow (top) and the class-conditional flow (bottom).}
\label{fig:nsf-architecture}
\end{figure}

A \textit{PLU linear transform} is a linear layer that is composed of three linear maps $\mathbf{W}=\mathbf{P}\mathbf{L}\mathbf{U}$, where each linear map is a $2d \times 2d$ matrix. $\mathbf{P}$ is a random permutation matrix that shuffles the $2d$ input dimensions, and $\mathbf{L}$ is a lower triangular matrix with ones on the diagonal. Similarly $\mathbf{U}$ is an upper triangular matrix. This guarantees that the composition $\mathbf{W}$ is invertible, and the inverse can be computed efficiently by solving two triangular systems, one for $\mathbf{U}$ and one for $\mathbf{L}$, then taking the inverse permutation $\mathbf{P}^{-1}$. We found that using PLU linear layers led to a noticeable improvement in both training and test likelihood.

Once the NSF is trained, sampling is performed by sampling $\mathbf{z} \sim p(\mathbf{z})$ then passing it through the flow $f_\theta$ to obtain a sample $\mathbf{x} = f_\theta(\mathbf{z})$.

\textbf{Training details}. Modulations are centered and normalized by elementwise mean \& standard deviation across the training modulation dataset, then scaled by a hyperparameter \verb!norm_factor! before they are fed into the NSF. For the base distribution of the flow, we use a Gaussian with 0 mean and standard deviation 0.25 (to match the inputs normalized by \verb!norm_factor!=4). We fix $p=3n_{bins} + 1$ where $n_{bins}=8$ is the number of bins for the rational quadratic spline bijector (\textit{cf.}\ \citet{durkan2019neural} for a detailed formulation of rational quadratic spline bijectors). See \autoref{fig:nsf-architecture} for the architectures of $g$ that we use for the unconditional flow and the class-conditional flow. We used Adam with a custom learning rate schedule that warms up linearly from 0 to 3e-4 for the first 4000 warmup iterations, then decays proportional to the square root of the iteration count. The linear layers in the conditioner all have output dimensionalities 128, and the label embeddings are 32-dimensional learnable parameters.

\textbf{Model selection}. The hyperparameters that we tuned are the number of flow layers, sweeped over $\{4,8,16,32,64,128,256,320\}$ and the dropout probability, sweeped over $\{0., 0.1, 0.2, 0.3\}$. The best model and checkpoint used for showing samples and performing inference was chosen by using test likelihood as the metric. See \autoref{tab:nsf-hyperparams} for selected hyperparameter values for each modulation dataset.

\textbf{Tips}. We found that for ShapeNet, data augmentation was key to prevent the NSF from overfitting.

\begin{table}[h!]
    \centering
    \small
    \begin{sc}
    \begin{tabular}{l|cccc}
        \toprule[1pt]
        Dataset & batch size & num flow layers & dropout probability & num iterations \\ \midrule
        ShapeNet Chairs (256 mod-dim) & 64 & 320 & 0.2 & 1e6 \\
        ShapeNet 10 Classes (256 mod-dim) & 64 & 128 & 0.3 & 1e6 \\
        ERA5 temperature (256 mod-dim) & 256 & 32 & 0.2 & 2e5 \\
        SRN Cars (512 mod-dim) & 64 & 8 & 0.3 & 1e5 \\
        \bottomrule[1pt]
    \end{tabular}
    \end{sc}
    \caption{Selected hyperparameter values for Neural Spline flows for each dataset.}
    \label{tab:nsf-hyperparams}
\end{table}

\subsection{Denoising Diffusion Probabilistic Models (DDPM)} \label{sec:ddpm}

\textbf{Background}. Diffusion models \cite{sohl2015deep} model the joint distribution of the data $\mathbf{x}_0$ (in our case the modulations) and noisy versions of the data $\mathbf{x}_1, \ldots, \mathbf{x}_T$ where $\mathbf{x}_t$ is obtained by linearly scaling $\mathbf{x}_{t-1}$ then adding Gaussian noise. This can be thought of as sequentially adding Gaussian noise to the data $\mathbf{x}_0$ to create variables $\mathbf{x}_1, \ldots, \mathbf{x}_T$, with suitably scaled noise such that the marginal distribution of $\mathbf{x}_T$ is a standard Gaussian. This is called the \textit{forward (diffusion) process} that is formulated as
\begin{equation}
    q(\mathbf{x}_{1:T}|\mathbf{x}_0) \coloneqq \prod_{t=1}^T q(\mathbf{x}_t|\mathbf{x}_{t-1}), \hspace{5mm} q(\mathbf{x}_t|\mathbf{x}_{t-1})  \coloneqq  \mathcal{N}(\mathbf{x}_t; \sqrt{1-\beta_t} \mathbf{x}_{t-1}, \beta_t \mathbf{I})
\end{equation}
for a fixed sequence of variances $\beta_{1:T}$. A desirable property of these Gaussian conditionals is that it admits sampling $\mathbf{x}_t$ directly from $\mathbf{x}_0$ without having to sample the intermediate steps since $q(\mathbf{x}_t|\mathbf{x}_0)$ is a Gaussian that can be written in closed form
\begin{equation}
    q(\mathbf{x}_t|\mathbf{x}_0) = \mathcal{N}(\mathbf{x}_t; \sqrt{\bar{\alpha}_t}\mathbf{x}_0, (1- \bar{\alpha}_t) \mathbf{I})
\end{equation}
where $\alpha_t \coloneqq 1 - \beta_t$ and $\bar{\alpha}_t \coloneqq \prod_{s=1}^t \alpha_s$. Note that this forward process is fixed and has no learnable parameters.

The aim of diffusion models is to learn the \textit{reverse (diffusion) process} that is formulated as:
\begin{equation}
p(\mathbf{x}_T) \coloneqq \mathcal{N}(\mathbf{x}_T; \mathbf{0}, \mathbf{I}), \hspace{3mm}
p_{\theta}(\mathbf{x}_{0:T}) \coloneqq p(\mathbf{x}_T)\prod_{t=1}^T p_{\theta}(\mathbf{x}_{t-1}|\mathbf{x}_t), \hspace{3mm}
p_{\theta}(\mathbf{x}_{t-1}|\mathbf{x}_t) \coloneqq \mathcal{N}(\mathbf{x}_{t-1}; \bm{\mu}_{\theta}(\mathbf{x}_t, t), \bm{\Sigma}_{\theta}(\mathbf{x}_t, t))
\end{equation}
The mean and variance of the reverse conditionals are parameterized by neural networks, whose parameters are optimized by minimizing
\begin{equation}
    \mathbb{E}_q \left[\text{KL}[q(\mathbf{x}_{t-1}|\mathbf{x}_t, \mathbf{x}_0)\| p_{\theta}(\mathbf{x}_{t-1}|\mathbf{x}_t)]\right]
\end{equation}
for each value of $t$, where a value of $t \in [2,1000]$ is randomly subsampled at each training iteration. It turns out that the term $q(\mathbf{x}_{t-1}|\mathbf{x}_t, \mathbf{x}_0)$ is also Gaussian and can be written in closed form, hence the above KL is tractable.

DDPM \cite{ho2020denoising} simplifies the above loss to the expression (see paper for derivation):
\begin{equation}
    \| \bm{\epsilon_t} - \bm{\epsilon}_{\theta}(\sqrt{\bar{\alpha}_t}\mathbf{x}_0 + \sqrt{1-\bar{\alpha}_t}\bm{\epsilon}_t, t) \|^2
\end{equation}
where $\mathbf{x}_t \sim q(\mathbf{x}_t|\mathbf{x}_0)$ is reparameterized as $\mathbf{x}_t = \sqrt{\bar{\alpha}_t}\mathbf{x}_0 + \sqrt(1- \bar{\alpha}_t) \bm{\epsilon}_t$ for $\bm{\epsilon}_t \sim \mathcal{N}(\mathbf{0}, \mathbf{I})$. Here $\bm{\epsilon}_{\theta}$ is the neural network predicting $\bm{\epsilon}_t$ from $\mathbf{x}_t$, which is a more convenient parameterization for optimizing the loss than explicitly parameterizing $\bm{\mu}_{\theta}, \bm{\Sigma}_{\theta}$.

Once trained, sampling is performed by sampling $\mathbf{x}_T \sim p(\mathbf{x}_T)$ then taking the reverse process $p_{\theta}(\mathbf{x}_{t-1}|\mathbf{x}_t)$ to sample $\mathbf{x}_{T-1}, \ldots, \mathbf{x}_0$ in sequence. The resulting $\mathbf{x}_0$ is a sample from the model.

\begin{figure}[t]
\begin{center}
\includegraphics[width=0.65\columnwidth]{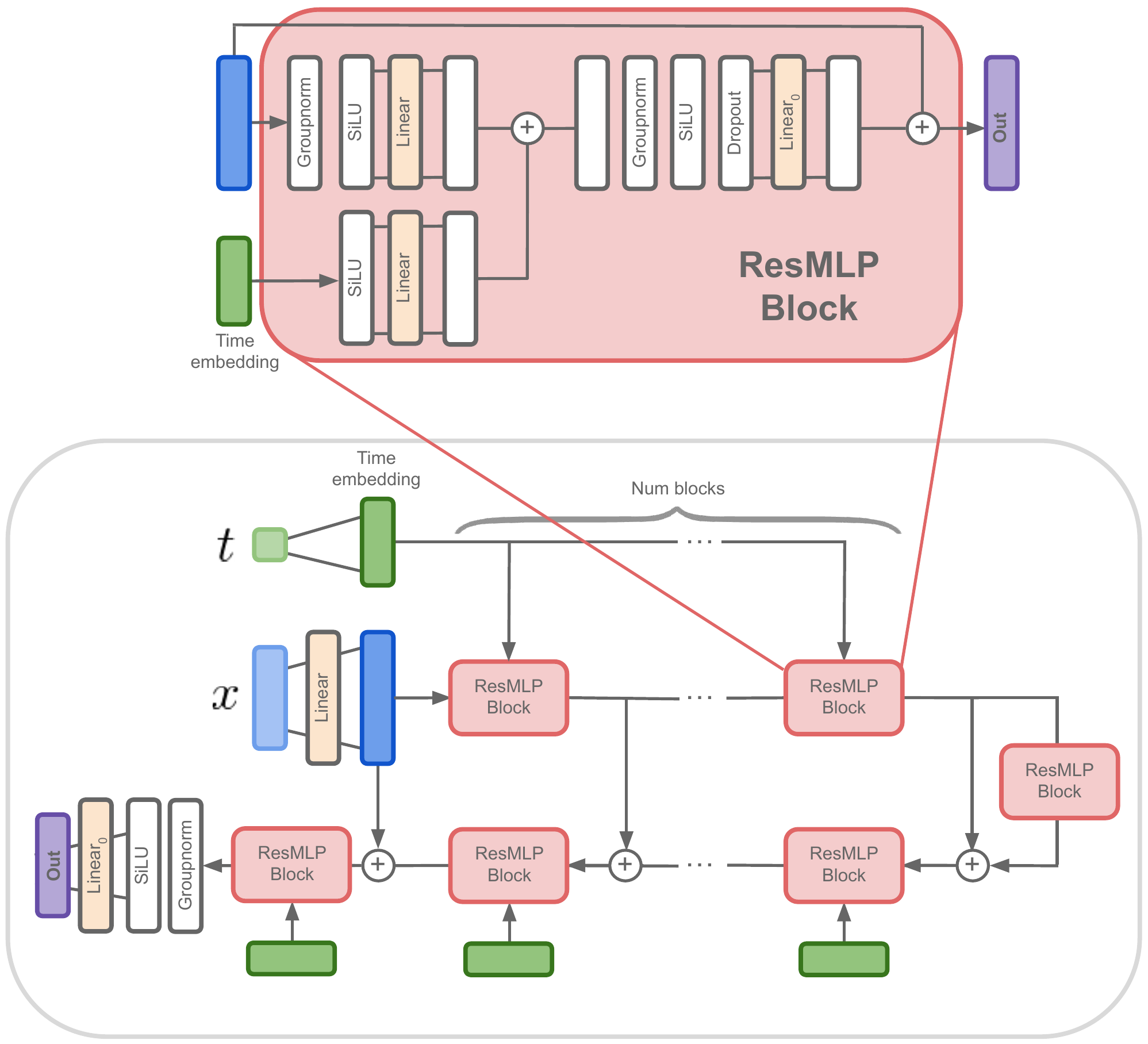}
\end{center}
\caption{Architecture for $\bm{\epsilon}_{\theta}$ in our implementation of DDPM. $\text{Linear}_0$ stands for a linear layer with zero initialization.}
\label{fig:ddpm-architecture}
\end{figure}

\textbf{Training details}.
As with NSF, modulations are centered and normalized by elementwise mean \& standard deviation across the training modulation dataset, and Adam is used with a custom learning rate schedule that warms up linearly from 0 to 3e-4 for the first 4000 warmup iterations, then decays proportional to the square root of the iteration count. We use $T=1000$ and $\beta_{1:T}=$\verb!np.linspace!$(10^{-4}, 0.02, T)$ following \citet{ho2020denoising}, and parameterize $\bm{\epsilon}_{\theta}$ as in \autoref{fig:ddpm-architecture}, which again closely follows the architecture used in \citet{ho2020denoising} except using ResidualMLPs in place of CNNs. 

\textbf{Model selection}. The hyperparameters that we tuned are the width of the ResidualMLPs, sweeped over $\{512,1024,2048,3072\}$, the number of blocks, sweeped over $\{4,8\}$ and the dropout probability in Reisudal MLPs, sweeped over $\{0., 0.1, 0.2, 0.3\}$. We take the model at the end of training to show samples, as we found that FID had limited correlation with perceptual quality (\textit{cf.}\ \Cref{sec:images}) and sample quality seemed to have converged by the end of training. For DDPM on 256-dimensional modulations of CelebA-HQ, we use a batch size of 128 with width 3072, 4 blocks and 0.3 dropout probability, training for 1e6 iterations. For DDPM on 64-dimensional modulations of SRN Cars, we use a batch size of 256 with width 512, 4 blocks and 0. dropout, training for 1e5 iterations.

\textbf{Tips.} 
We found that width was the single most important hyperparameter, where wider models gave better samples. Also dropout was important to prevent the model from memorizing the training set.

\subsection{Neural Radiance Fields (NeRF)} \label{sec:nerf-details}
\textbf{Background}. In NeRF \cite{mildenhall2020nerf}, a 3D scene is expressed by a (scene) function that maps a 3D point's coordinates and viewing direction of the camera looking at that point to the RGB colour of the point and its density. 
In practice this function is an MLP (LatentModulatedSIREN in our case), that is trained on a dataset of multiple views of the same scene, using each view's pose information and its focal length, as described in \Cref{sec:datasets}.
Given this function, the model uses a volumetric rendering formula to construct views of the scene, by shooting rays from the camera to each pixel and aggregating the function values at points along each ray in a differentiable manner. The MLP parameters of the scene function are optimized by minimizing the reconstruction loss on all views. See \citet{mildenhall2020nerf} for details.

\textbf{Training details}. For simplicity, we use a minimal volumetric rendering algorithm (see \autoref{lst:nerf-rendering}) that closely follows the implementation of \citet{tancik2021learned}, a simplified version of the rendering used in \citet{mildenhall2020nerf}. We do not use hierarchical volume sampling (coarse/fine points on ray), use a single SIREN for outputting the RGB \& density (as opposed to the standard practice of having separate networks/heads for each), no view dependence (usually the viewing direction is fed into the scene function as extra information). The (near, far) values that determine the start and end points for sampling on the ray are fixed to (0.8, 1.8).

\textbf{Model selection}. Note that the array containing all training views is large ($50$ views each with 128$\times$128$\times$3 pixel values), hence we can only use a single scene for each training iteration per device (batch size=1 per device, so must use multiple devices for larger batch sizes). Further we must subsample views and pixels over which we optimize the reconstruction error, in order to fit the meta-learning in memory. Hence for meta-learning we tune the hyperparameters such as number of points per ray, sweeped over (16, 32, 64), the number of subsampled views, sweeped over (4, 8, 16), and the number of subsampled pixels, sweeped over (512, 1024, 2048). The optimal hyperparameters for the 512-dimensional modulation dataset is: 32 points per ray, 16 views and 512 pixels per view. Note that for the modulation dataset creation, we can use more views and pixels as we only need to run the inner loop of the meta-learning (rather than having to backpropagating through it, which is memory costly). Hence we used 32 views and 2048 pixels per view for the modulation dataset creation.

\textbf{Tips}. We found it best to max out memory by trying different combinations of the 3 hyperparamters (number of points per ray, number of subsampled views, number of subsampled pixels)

% \textbf{Side Note.} We expect using a more sophisticated rendering algorithm will help improve reconstruction accuracy of modulations in our framework. Moreover advances in scaling NeRF such as \citet{sitzmann2021light} that removes the need ray-marching will help reduce memory costs and allow us to further scale up our method.

\definecolor{codegreen}{rgb}{0,0.6,0}
\definecolor{codegray}{rgb}{0.5,0.5,0.5}
\definecolor{codepurple}{rgb}{0.58,0,0.82}
\definecolor{backcolour}{rgb}{0.95,0.95,0.92}

\lstdefinestyle{mystyle}{
    backgroundcolor=\color{backcolour},   
    commentstyle=\color{codegreen},
    keywordstyle=\color{magenta},
    numberstyle=\tiny\color{codegray},
    stringstyle=\color{codepurple},
    basicstyle=\ttfamily\footnotesize,
    breakatwhitespace=false,         
    breaklines=true,                 
    captionpos=b,                    
    keepspaces=true,                 
    numbers=left,                    
    numbersep=5pt,                  
    showspaces=false,                
    showstringspaces=false,
    showtabs=false,                  
    tabsize=2
}

\lstset{style=mystyle}
\newpage
\begin{lstlisting}[language=Python, caption=Minimal volumetric rendering, label={lst:nerf-rendering}, numbers=none]
import jax
import jax.numpy as jnp

def render_pose(model, params, height, width, focal, pose,
                num_points_per_ray, near, far, white_background):
  """Render view of a single scene.
  
  Args:
    model: MLP with input_size = 3, output_size = 4 (3 for RGB and 1 for density.)
    params: Model params.
    height: Height of image.
    width: Width of image.
    focal: Focal length.
    pose: Pose (camera to world matrix) of shape (4, 4).
    num_points_per_ray: Number of points per ray. Splits rays
        into equally spaced points.
    near: Point nearest to the camera where ray starts.
    far: Point furthest from the camera where ray ends.
    white_background: If True sets default RGB value to be 1,
        otherwise will be set to 0 (black).
  Returns:
    rgb_map: Rendered view. Array of shape (..., 3).
  """
  # Compute rays and split into ray origins and ray directions
  i, j = jnp.meshgrid(jnp.arange(width), jnp.arange(height), indexing='xy')
  dirs = jnp.stack([(i - width * .5) / focal,
                    -(j - height * .5) / focal,
                    -jnp.ones_like(i)], -1)
  rays_d = jnp.sum(dirs[..., jnp.newaxis, :] * pose[:3, :3], -1) # (..., 3)
  rays_o = jnp.broadcast_to(pose[:3, -1], rays_d.shape)  # (..., 3)
  # Compute 3D query coordinates
  z_vals = jnp.linspace(near, far, num_points_per_ray)
  # The below line uses (a lot of) broadcasting. In terms of shapes:
  # (...,1,3) + (...,1,3) * (num_points_per_ray,1) = (...,num_points_per_ray,3)
  coords = rays_o[..., None, :] + rays_d[..., None, :] * z_vals[..., :, None]
  out = model(params, coords)  # (..., num_points_per_ray, 4)
  # Compute colors and volume densities
  rgb, density = out[..., :3], out[
      ..., 3]  # (..., num_points_per_ray, 3), (..., num_points_per_ray)
  density = jax.nn.elu(density, alpha=0.1) + 0.1 # Ensure density is positive
  density = jnp.clip(density, 0., 10.)  # upper bound density at 10
  # Calculate distance between consecutive points along ray. 
  distance_between_points = z_vals[..., 1:] - z_vals[..., :-1]
  distances = jnp.concatenate([distance_between_points,
                               1e-3 * jnp.ones(1)])  # (num_points_per_ray,)
  # Alpha will have a value between 0 and 1
  alpha = 1. - jnp.exp(-density * distances)  # (..., num_points_per_ray)
  # Ensure transmittance is <= 1 (and greater than 1e-10)
  trans = jnp.minimum(1., 1. - alpha + 1e-10)
  # Make the first transmittance value along the ray equal to 1 for every ray
  trans = jnp.concatenate([jnp.ones_like(trans[..., :1]), trans[..., :-1]],
                          -1)  # (..., num_points_per_ray)
  cum_trans = jnp.cumprod(trans, -1)  # T_i in Equation (3) of NeRF paper.
  weights = alpha * cum_trans  # (..., num_points_per_ray)
  # Sum RGB values along the ray
  rgb_map = jnp.sum(weights[..., None] * rgb, -2)  # (..., 3)
  # Optionally make background white
  if white_background:
    acc_map = jnp.sum(weights, -1)  # Accumulate weights   (...)
    rgb_map = rgb_map + (1. - acc_map[..., None])  # Add white background
  return rgb_map
\end{lstlisting}

\subsection{Classification} \label{sec:classification-details}
We use the default MLP in Haiku \cite{haiku2020github} with SiLU/swish activations \cite{hendrycks2016gaussian, ramachandran2017searching} and dropout for classifying modulations. The width is sweeped over (128, 256, 512) and the depth over (2,3,4).

For the 3D CNN baseline we closely follow the architecture in \cite{maturana2015voxnet}, which applies a sequence of 3D Conv layers with stride 2, followed by a flattening, then 2 linear layers, the first of which has width 128 and the last of which outputs logits for the classification. Each 3D Conv/linear layer is followed by a SiLU activation then dropout. 
The 3D Conv layers all have the same number of output channels sweeped over (16, 32, 64), and the number of 3D Conv layers is also sweeped over (4,5,6).

Both classifiers are optimized for 3e4 iterations with Adam at a fixed learning rate of 1e-4, and a total batch size of 1024. For the 3D CNN, this is spread over 8 devices to fit the batch in memory. In \Cref{sec:classification}, we show results for the best-performing MLP with 128 width and the best-performing 3D CNN with 16 output channels, with error bars calculated over 3 different random seeds. In \Cref{sec:further-experiments} we show results for larger models.

\subsection{FID scores} \label{sec:fid}

For CelebA-HQ, we evaluated our model in terms of FID \cite{heusel2017gans} but found that it was not a meaningful perceptual metric for our model. FID scores at the latter stages of training were sometimes worse than early in training even though samples were perceptually of much higher quality. Further, the FID score of the functaset itself (fitted directly to the dataset) is 28.4 for 256-dimensional modulations and 17.2 for 512-dimensional modulations, despite both being perceptually very close to the original dataset (\textit{cf.}\ \autoref{fig:orig-vs-rec}). We believe this is due to FID's property of over-penalizing blurriness, with very similar effects observed by \citet{du2021gem} and \citet{razavi2019generating}. Nonetheless, our model obtains an FID score of 40.4, compared to 13.5 for GASP, 5.9 for StyleGANv2 \cite{karras2020analyzing} and 175.3 for a VAE \cite{du2021gem}. 

\subsection{Figure 2 details} \label{sec:figure2}

The original $128^3$ voxel grid was downscaled using the \verb!scipy.ndimage.zoom! function to each of the resolutions $8^3$, $16^3$, $32^3$, $64^3$, $128^3$. For each grid, we performed a manual search over SIREN (\textit{cf.}\ \Cref{sec:siren-modulation}) architectures (width, depth and $\omega_0$) to find the smallest architecture that could fit the voxel grid to 99.9\% accuracy. The settings for each architecture are summarized in \autoref{tab:fig2}.

\begin{table}[h!]
    \centering
    \small
    \begin{sc}
    \begin{tabular}{c|ccccc}
        \toprule[1pt]
        resolution & depth & width & $\omega_0$ & $n_{\text{params}}$\\ \midrule
        $8^3$    & 4  & 6  & 8  & 157 \\
        $16^3$   & 4  & 8  & 12 & 257 \\
        $32^3$   & 5  & 12 & 12 & 685 \\
        $64^3$   & 7  & 20 & 14 & 2621 \\
        $128^3$  & 10 & 32 & 50 & 9665 \\
        \bottomrule[1pt]
    \end{tabular}
    \end{sc}
    \caption{Hyperparameter values used for \autoref{fig:curse-of-disc}.}
    \label{tab:fig2}
\end{table}

\section{Modulation analysis}\label{sec:modulation-analysis}

\subsection{Perturbation analysis} \label{sec:perturbation-analysis}

\begin{figure}[t]
\begin{center}
\includegraphics[width=0.75\columnwidth]{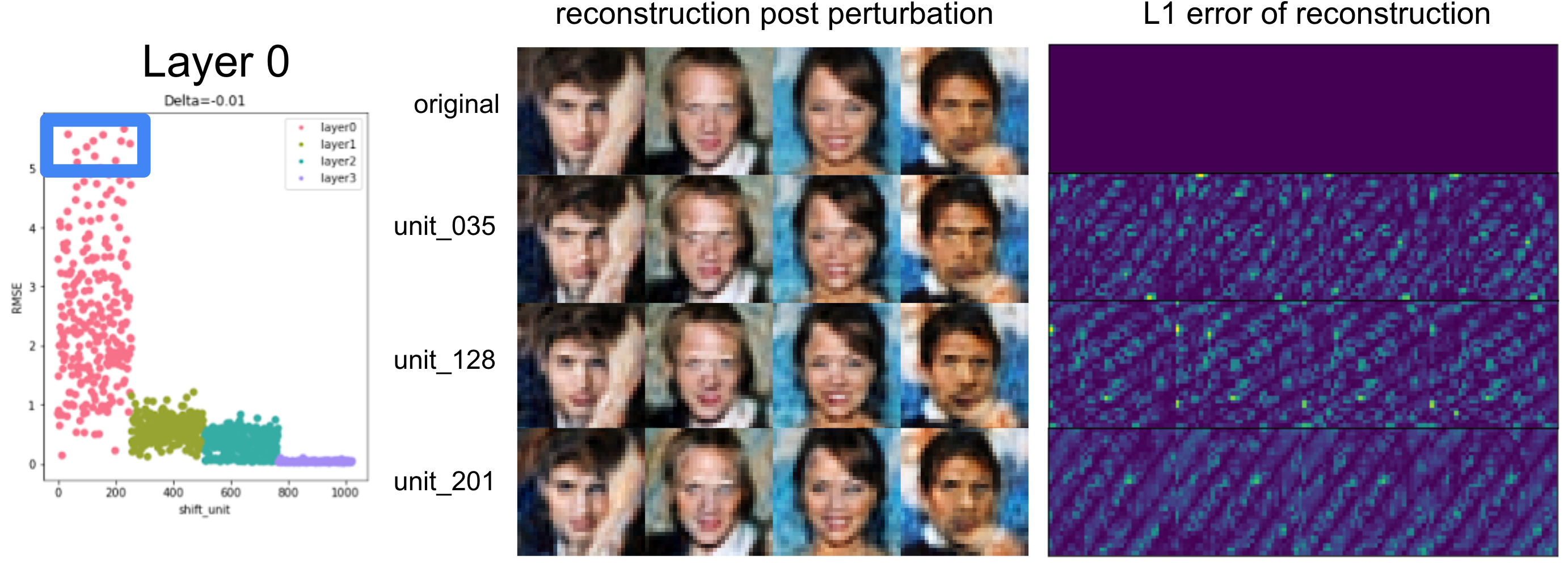}
\includegraphics[width=0.75\columnwidth]{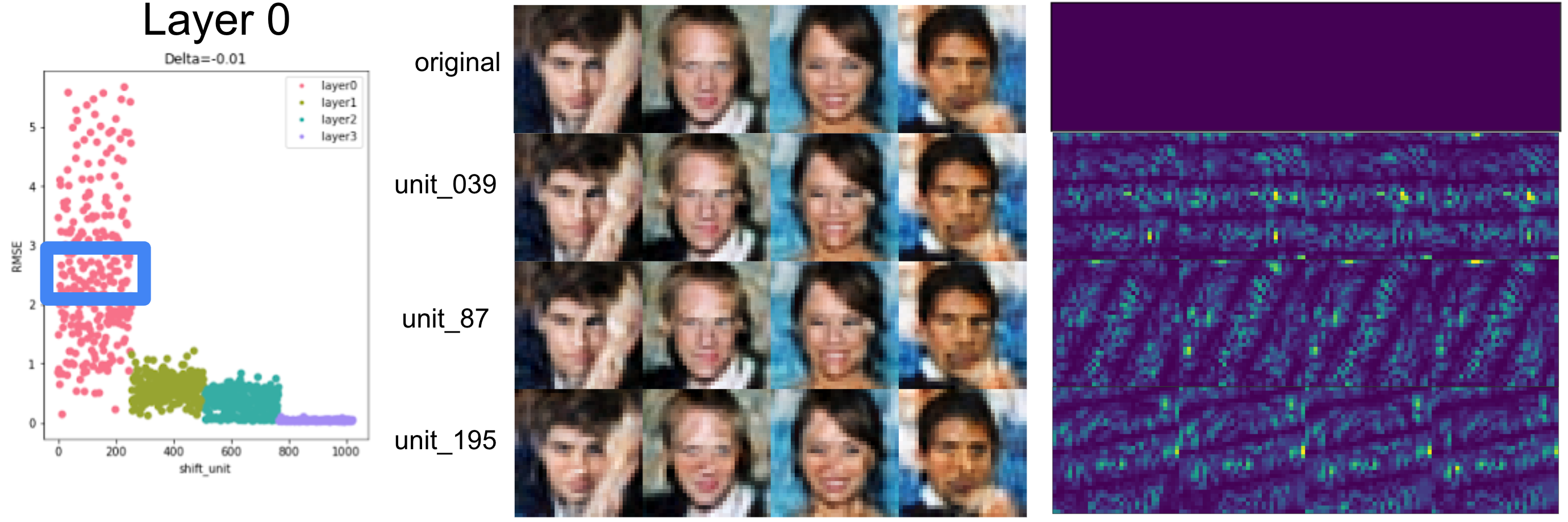}
\includegraphics[width=0.75\columnwidth]{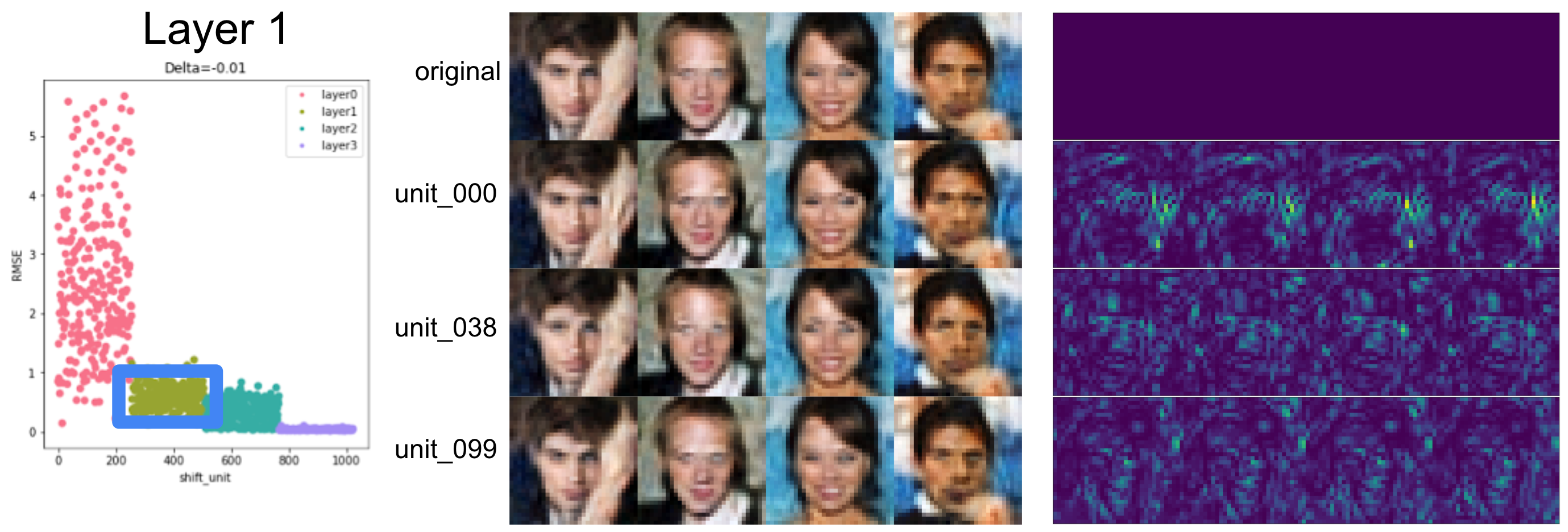}
\includegraphics[width=0.75\columnwidth]{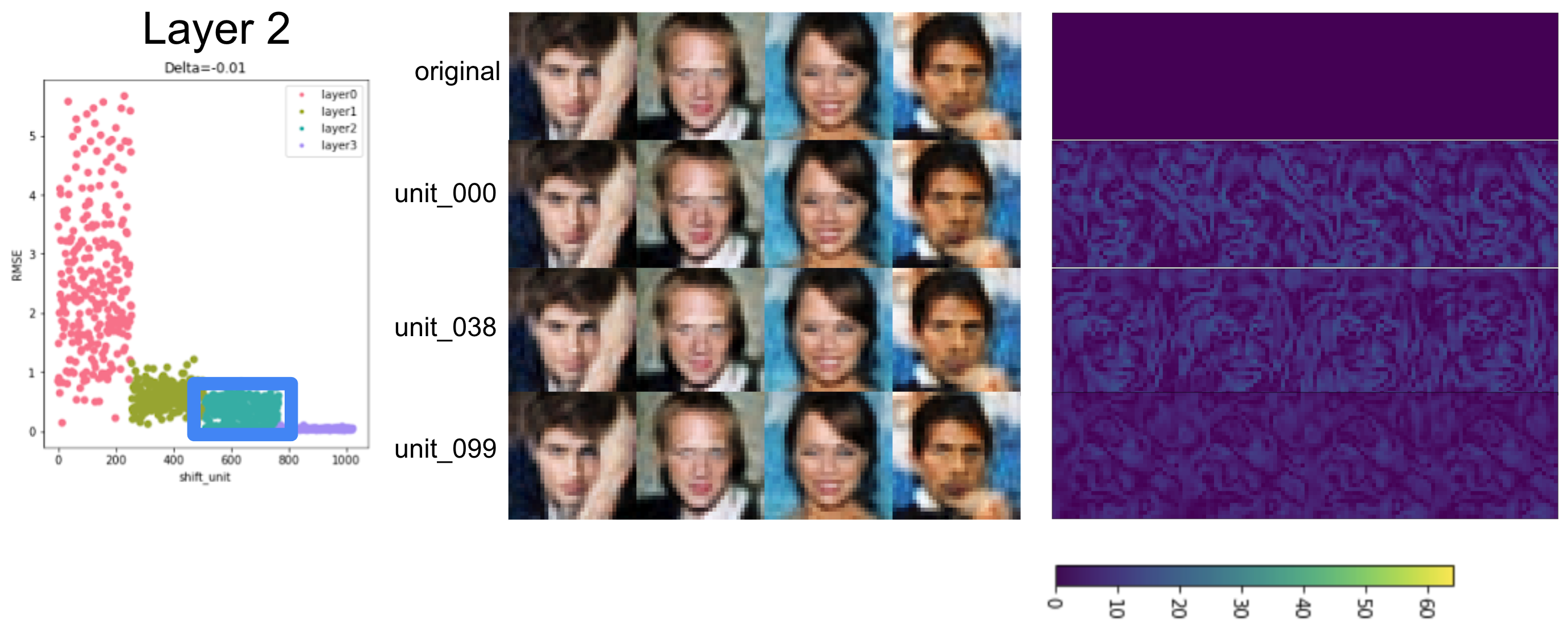}
\end{center}
\caption{Visualization of errors in pixel space when perturbing individual modulation dimensions of each layer by 0.01.}
\label{fig:perturbation-analysis}
\end{figure}

We provide intuition for the role of each shift modulation dimension by perturbation analysis of individual shift modulation dimensions in \autoref{fig:perturbation-analysis}, for functa with 4-layer SIREN modulations with width 256 trained on the 32$\times$32 CelebA-HQ training set.
The left column shows a plot of modulation index ($x$-axis) against RMSE in the perturbed reconstruction ($y$-axis) after applying a perturbation of $\delta=-0.01$ to the scalar shift modulation at that index. We colour code modulation units by layer. For each figure row we take 4 different shift modulation units from a given layer with similar RMSEs (sampled from the blue box) and visualize reconstructions after perturbing each unit on 4 different training images (middle column), along with the L1 error of the reconstruction for each spatial dimension (right column).

The first row shows perturbations of 4 units of the SIREN's initial hidden layer (layer 0) that have high RMSE ($>5$).
We see that these units are responsible for features that are roughly periodic along the diagonal, with different periodicity and direction - the troughs show zero error along the diagonal lines.

The second row shows 4 units of medium RMSE ($\in [2,3]$), where the periodicity remains but the frequency is lower than the high RMSE units. Hence it seems that the initial hidden layer contains a range of units that account for features of different frequencies in pixel space.

The third row shows 4 units in the next layer (layer 1), where the errors in pixel space are no longer periodic, and seem to be aligned with the edges of the image. Also the magnitudes of the perturbation in the image space are smaller than the previous layer, consistent with their RMSE values. 

The final row shows 4 units in the subsesquent layer (layer 2) again showing errors that seem to be aligned with the edges, but are a bit smoother and the magnitudes of the perturbation in the image space are yet smaller than previous layers.

Overall RMSE due to perturbation tends to decrease with depth of the SIREN, indicating that the modulations of earlier layers have a greater role in modelling the variations across the dataset compared to later layers.

One other interesting observation from the right column is that the perturbations for a given modulation unit lead to very similar error patterns across all images. This suggests that we can think of each modulation dim as representing the coefficient of a basis function, since changing the value of this coefficient leads to similar changes for all images. However note from the figure that these basis functions are non-local. This might pose challenges for downstream neural networks to extract high level information from such modulations, given that locality is an important inductive bias for many architectures such as CNNs and Transformers. Hence we may wish to explore other decoder architectures such as Gaussian activations \cite{ramasinghe2021beyond} or Multiplicative Gabor Networks \cite{fathony2020multiplicative} whose modulations may correspond to more localised information.

\subsection{Alternative Modulation architectures} \label{sec:modulation-architectures}

We explored alternative forms of modulations, which we found to be similar if not worse than latent modulations introduced in \Cref{sec:modulations}:

\textbf{Scale+Shift, Scale-only}. We only used shift modulations which are added to the SIREN activations. We found that when using both scale and shift, the scale values are almost always optimized to 1, hence redundant, giving similar performance to shift-only modulations. Scale-only modulations gave noticeably worse performance, resulting in around 5dB drop in reconstruction PSNR on images. We conjecture that this is due to gradients with respect to shift being independent to activation magnitude, whereas gradients with respect to scale is linearly dependent, resulting in higher magnitude gradients for shift modulations that allow faster optimization.

\textbf{SubsetModulatedSiren}. As shown in \Cref{sec:perturbation-analysis}, reconstructions are more sensitive to earlier modulation layers than later ones. Hence we can reduce the number of shift modulations by only using them for the first few layers of the MLP corresponding to the INR. While also competitive, we found that this approach does slightly worse than using latent modulations. Although we haven't yet tried, we expect better performance when these two approaches are combined, so that the latent code is only mapped to a subset of modulations.

\textbf{ModSine}. We experimented with the architecture in \citet{mehta2021modulated}. Instead of mapping a latent vector to the modulations, we map a latent vector through a network that has the same shape as the base network and treat the inner layers of the mapping network as modulations. However, we found that this approach gave lower reconstruction accuracy than latent modulations for the same modulation dimension.

\textbf{Concatenating latent to input of SIREN or ReluMLP+pos-enc} We have also tried meta-learning with the architecture in \citet{sitzmann2019metasdf} that concatenates the latent vector as input to the MLP decoder, and results were worse than latent modulations: we tried both SIREN and ReluMLP+pos-enc, sweeping over optimization hyperparameters and controlling for latent \& model size. The best was a SIREN with concatenated latent, giving test PSNR 23.0dB for latent-dim=256 on CelebA, notably worse than latent modulation's 25.6dB test PSNR in \autoref{tab:mod-rec-acc}.

\section{Negative Results} \label{sec:didnt-work}

\begin{figure}[h!]
\begin{center}
\includegraphics[width=0.5\columnwidth]{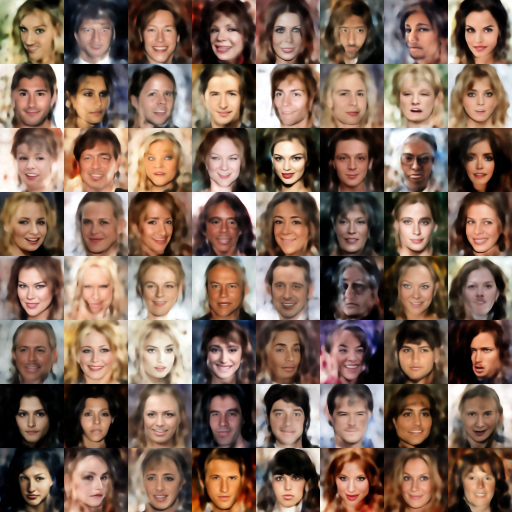}
\end{center}
\caption{Uncurated samples from autoregressive Transformer trained on 256-dim modulations for CelebA-HQ 64$\times$64.}
\label{fig:celeba-transformer-samples-256}
\end{figure}

\begin{figure}[h!]
\hspace{112pt} Init \hspace{24pt} Step 1 \hspace{18pt} Step 2 \hspace{18pt} Step 3 \hspace{18pt} Step 4 \hspace{18pt} Target
\begin{center}
\includegraphics[width=0.59\columnwidth]{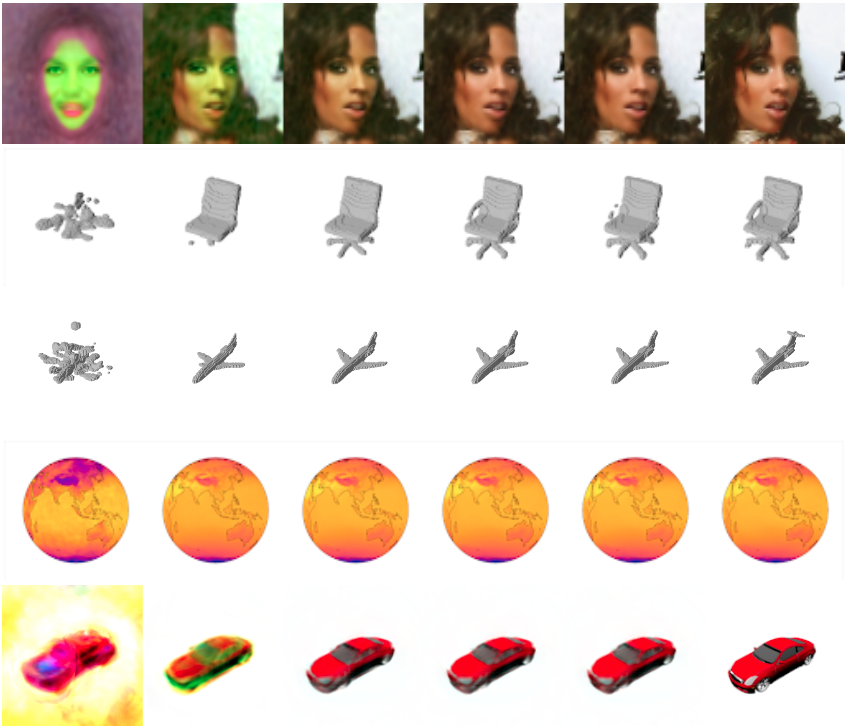}
\end{center}
\caption{Same as \autoref{fig:meta-learning-visualization} but with an extra inner loop step, and extra row for ShapeNet 10 classes.}
\label{fig:meta-learning-visualization-4step}
\vspace{-5mm}
\end{figure}

\textbf{Autoregressive Transformers}. As a third type of generative model, we tried training Transformers \cite{vaswani2017attention} on modulations (after discretizing each modulation to 5 bits after which we saw little loss in reconstruction accuracy). Yet the resulting samples shown in \autoref{fig:celeba-transformer-samples-256} were clearly perceptually worse than diffusion. We suspect this is because it is difficult to impose a meaningful ordering on the modulations for autoregressive modelling - the ordering is something we haven't played around with too much.

\textbf{Attention within diffusion and flow}. For the NSF, we tried using Tranfsormers for the conditioner but this did not outperform the MLP. For diffusion, we also tried using Transformers in place of residual MLPs, but this also performed worse.

\textbf{First order MAML}. We also tried first order meta-learning that does not backpropagate through the inner loop, to see if we could save memory with little sacrifice in performance. However this led to a drastic reduction in reconstruction PSNR, on the order of 10dB for images.

\textbf{Using more than 3 gradient steps for fitting modulations}. \autoref{fig:meta-learning-visualization-4step} shows reconstructions with up to 4 steps. Note that in most cases the difference between the 3rd and 4th steps is imperceptible, and for chairs the performance actually deteriorates (likely due to the fixed learning rate for the inner loop being too high). Using a greater number of steps, e.g.\ 50, we can obtain perceptibly better reconstructions for images (around 2dB better) but the modulation space is now less smooth (more gradient steps away from the initial modulation), so samples from generative modelling showed worse sample quality.

\textbf{Using more than 3 inner loop steps for meta-learning}. For scene data, we tried meta-learning with 4 or 5 inner loop steps. This led to a larger memory consumption, and hence we had to subsample fewer views and pixels per scene, yielding no noticeable improvement in terms of PSNR. For scenes, it seems very likely that the minimal rendering is the bottleneck for our setup.

\textbf{Flow on 256 mod-dim for images} performed noticeably worse than DDPM in terms of sample quality for images.

\textbf{Flow on 512-dimensional modulations for voxels} gave much worse sample quality than 256-dimensional modulations. We conjecture that using fewer modulations helps obtain a smoother modulation space, despite having worse PSNR/voxel accuracy.

\textbf{Diffusion on 256 mod-dim for voxels} was also noticeably worse than flows. Overall it appears as though flows are more suited for ShapeNet and diffusion for CelebA-HQ, and we conjecture that this is a property of the dataset rather than the modality.

\textbf{Bounding box subsampling for SRN Cars}. For scenes, when subsampling pixels from views for meta-learning, we tried subsampling pixels within the bounding box that contains the car, to avoid sampling white background pixels that are less informative than the foreground car pixels. However this didn't work better than subsampling at random, and we found that subsample white background pixels helps to achieve higher PSNR to accurately recover the white background.

\textbf{Auto-decoder approach for learning per-datapoint modulations}. Encouraged by the success of \citet{park2019deepsdf} on learning signed-distance functions, we have tried the auto-decoder approach for learning per-datapoint modulations on images. In particular we store a randomly initialized set of (latent) modulations per image (unshared params), along with a set of shared parameters that we jointly optimize by minimizing reconstruction error on each batch during training. This is attractive because it does not require memory-expensive double loop optimizations like meta-learning.
However for auto-decoding, training was unstable and PSNR was notably worse than for meta-learning, despite carefully tuning separate learning rates for the shared and unshared parameters with a varying number of gradient updates for each set of parameters. It is unclear as to why this approach seems to work for signed-distance functions yet does not seem to work for other data modalities.

\section{Further Results} \label{sec:further-experiments}

\begin{figure}[h!]
\begin{center}
\includegraphics[width=\columnwidth]{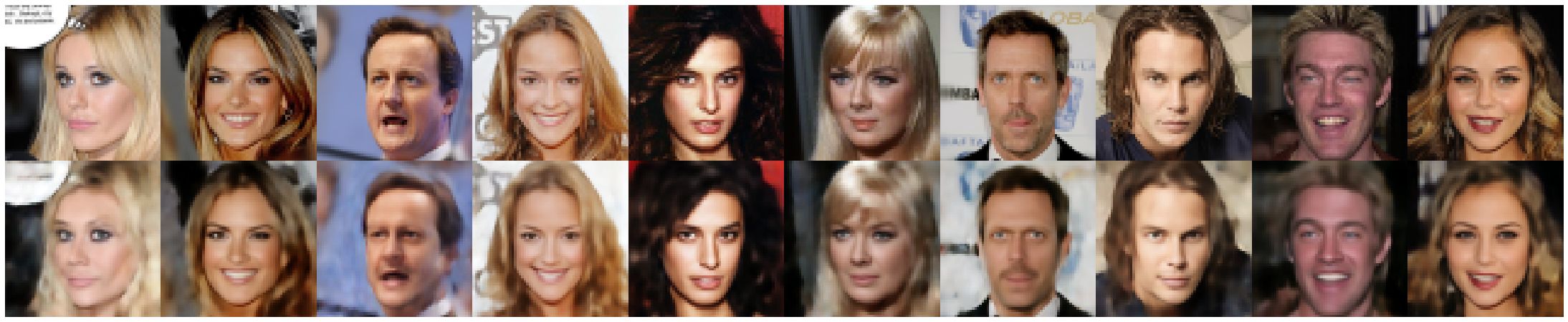}
\includegraphics[width=\columnwidth]{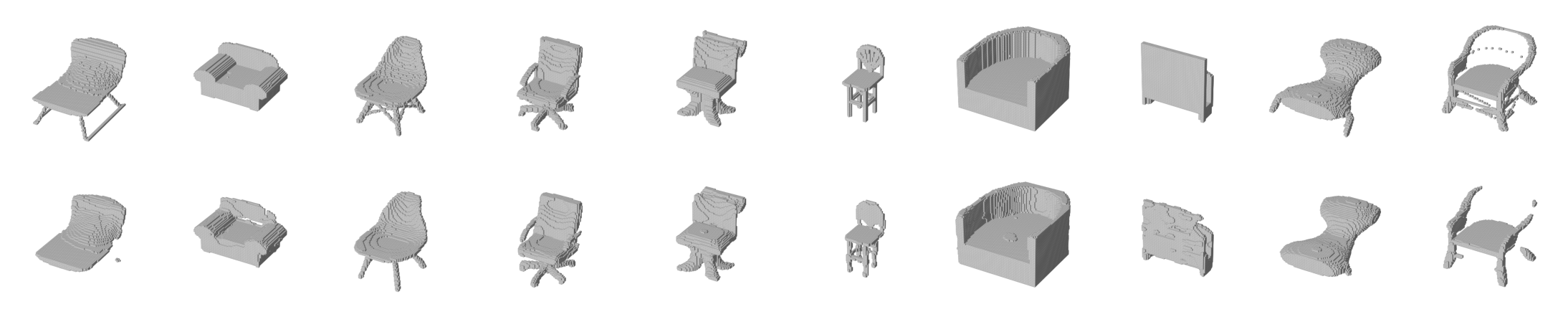}
\includegraphics[width=\columnwidth]{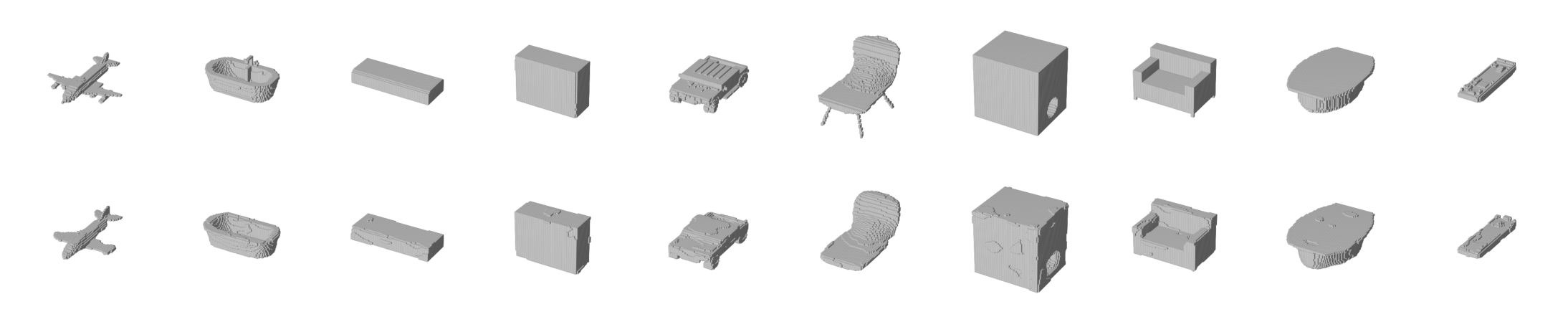}
\includegraphics[width=\columnwidth]{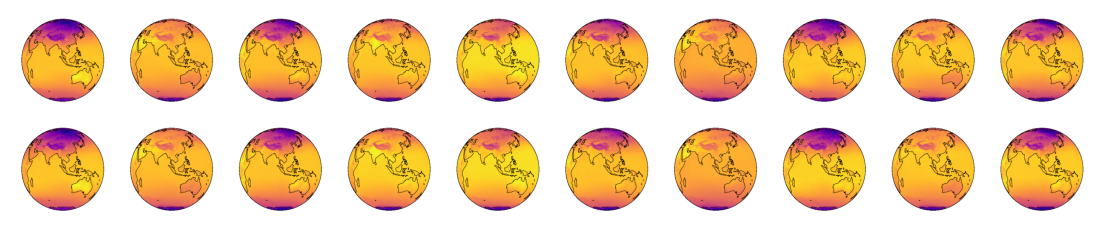}
\includegraphics[width=\columnwidth]{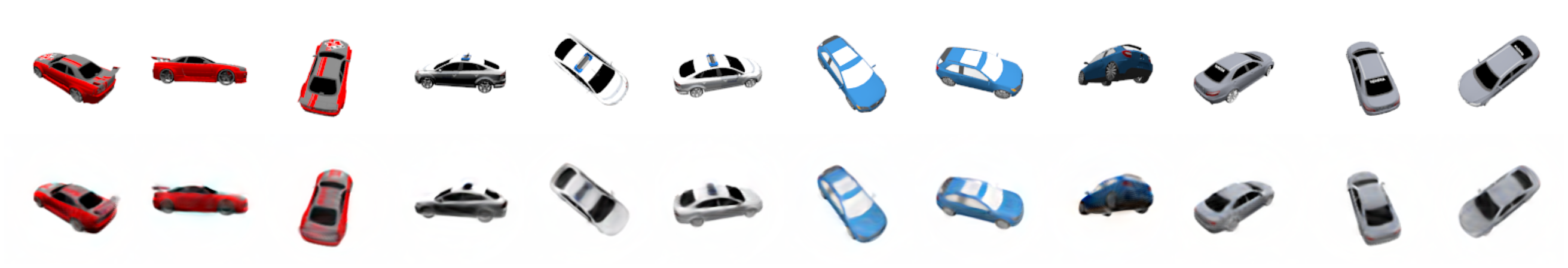}
\end{center}
\caption{Original vs reconstruction for 256-dimensional modulations on the first few data points of each training dataset.}
\label{fig:orig-vs-rec}
\end{figure}

\textbf{Original vs Reconstructions from modulations}. In \autoref{fig:orig-vs-rec} we compare the original array dataset (top row) against the reconstruction from 256-dimensional modulations obtained from meta-learning for each of the data modalities (used to compute the middle column for \autoref{tab:mod-rec-acc}). We see that the reconstructions are reasonably close to the original, although some fine detail can be missed for complex shapes and scenes. As mentioned in \Cref{sec:didnt-work}, using more inner loop gradient steps to fit the modulations can result in better reconstructions but at the sacrifice of performance for the downstream task e.g.\ generative modelling.

\begin{figure}[h!]
\begin{center}
\includegraphics[width=0.5\columnwidth]{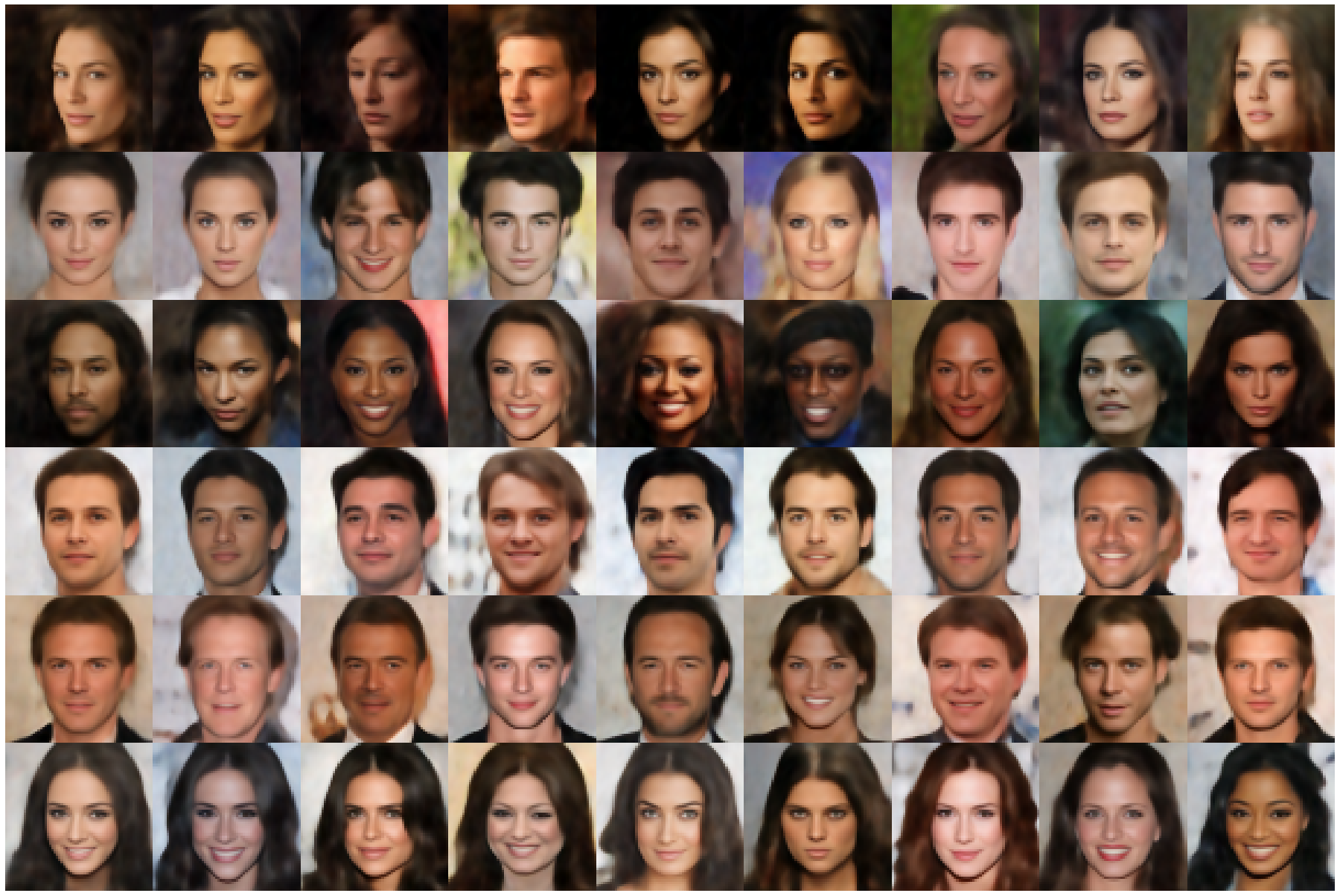}
\end{center}
\caption{Nearest neighbours of samples from diffusion model trained on 256-dim CelebA-HQ 64$\times$64 modulations. Leftmost column are samples, and subsequent columns are the closest neighbours in L2 distance in modulation space.}
\label{fig:celeba-diffusion-samples-256-knn}
\end{figure}

\textbf{Image samples and nearest neighbours in training set}. In \autoref{fig:celeba-diffusion-samples-256-knn} we show the nearest neighbours of samples from the diffusion model trained on 256-dimensional CelebA-HQ modulations. As can be seen, the samples are reasonably different to the training data, indicating that the diffusion hasn't simply memorised the training set. We show similar results for diffusion models trained on 512-dimensional modulations.

\begin{figure}[h!]
\begin{center}
\includegraphics[width=0.5\columnwidth]{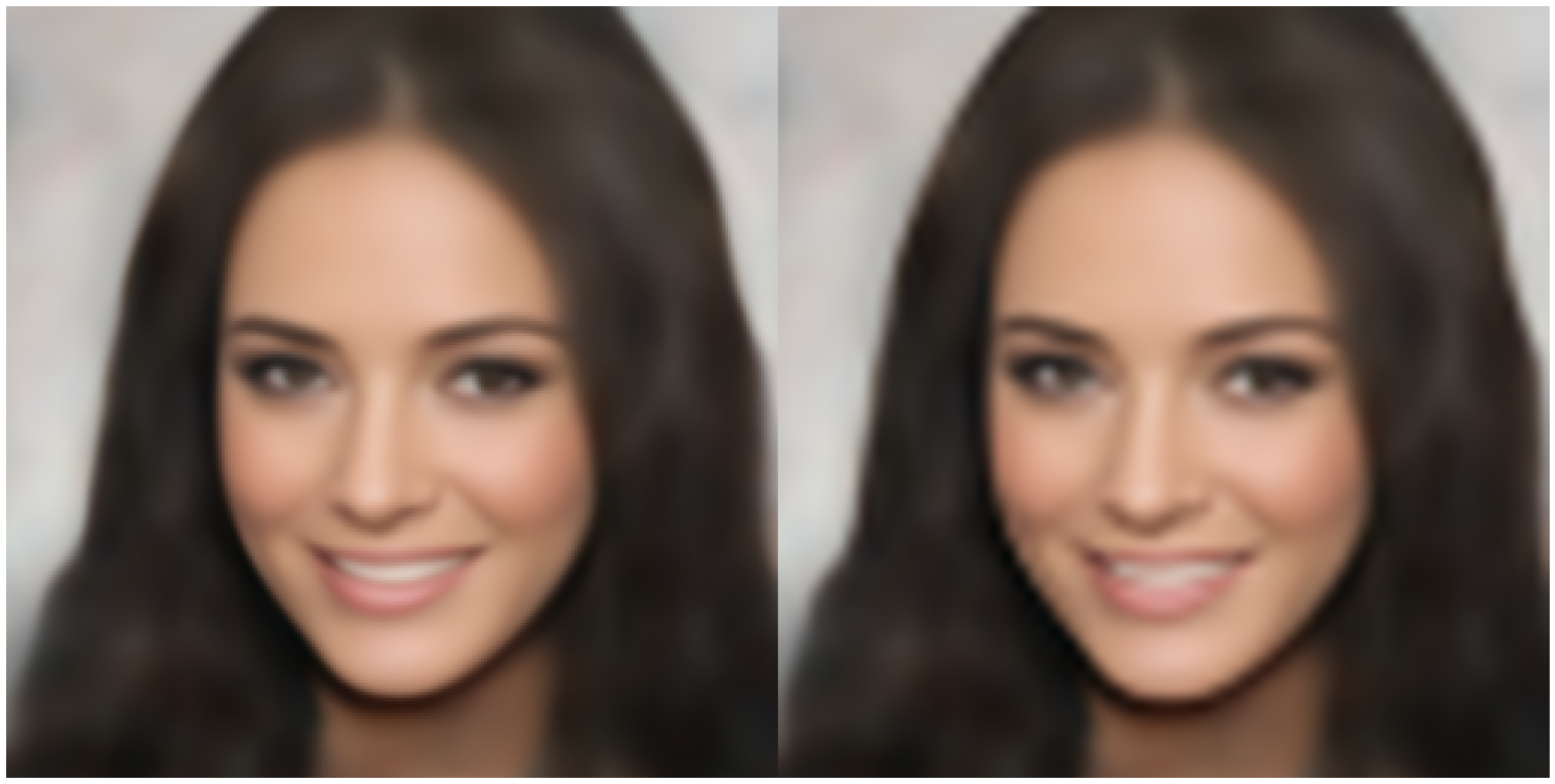}
\end{center}
\caption{Comparison of rendering sample at 512 resolution vs rendering at 64 resolution then upsampling to 512 by bicubic interpolation}
\label{fig:celeba-diffusion-sample-high-res-bicubic-interpolation-comparison}
\end{figure}

\textbf{High resolution samples}. In addition to unconditional generation, we also use our model to generate high resolution samples ($512\times512$) as shown in \autoref{fig:celeba-diffusion-sample-high-res-bicubic-interpolation-comparison}.
The left is obtained by querying the function represented by a 256 modulation sampled from the flow on a $512\times512$ grid, and the right is obtained by querying the same function on a $64\times64$ grid then applying bicubic interpolation to upscale to $512\times512$ resolution. We can see that the left is more crisp than the right, especially around the jawline and the teeth, indicating that the flow has learned a good internal representation of faces from various images in the CelebA-HQ dataset.

\begin{figure}[h!]
\begin{center}
\includegraphics[width=\columnwidth]{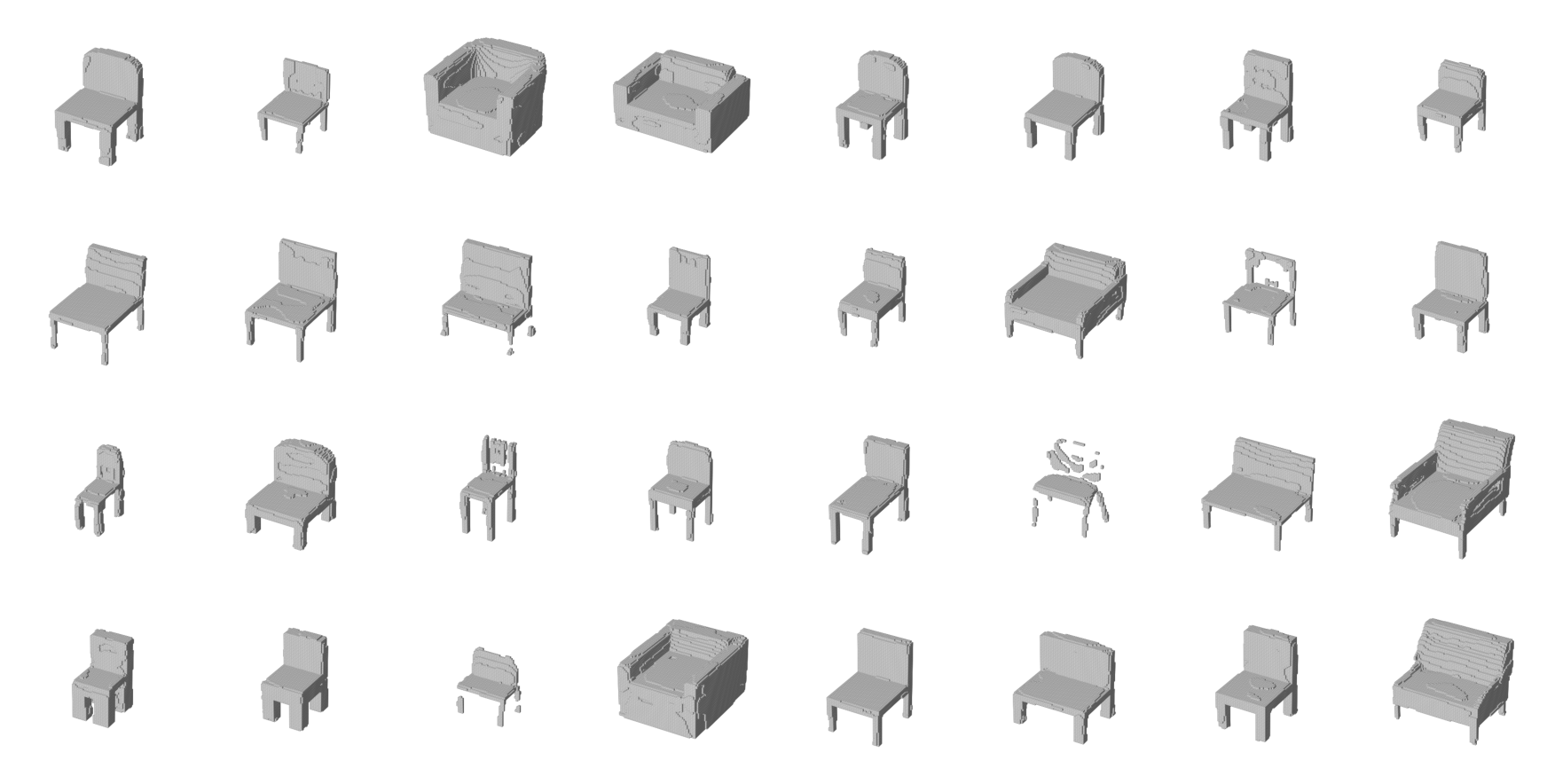}
\end{center}
\caption{Uncurated samples from flow trained on 256-dim ShapeNet Chairs $64^3$ modulations, temperature 0.95.}
\label{fig:chair-flow-samples-256}
\end{figure}

\begin{figure}[h!]
\begin{center}
\includegraphics[width=\columnwidth]{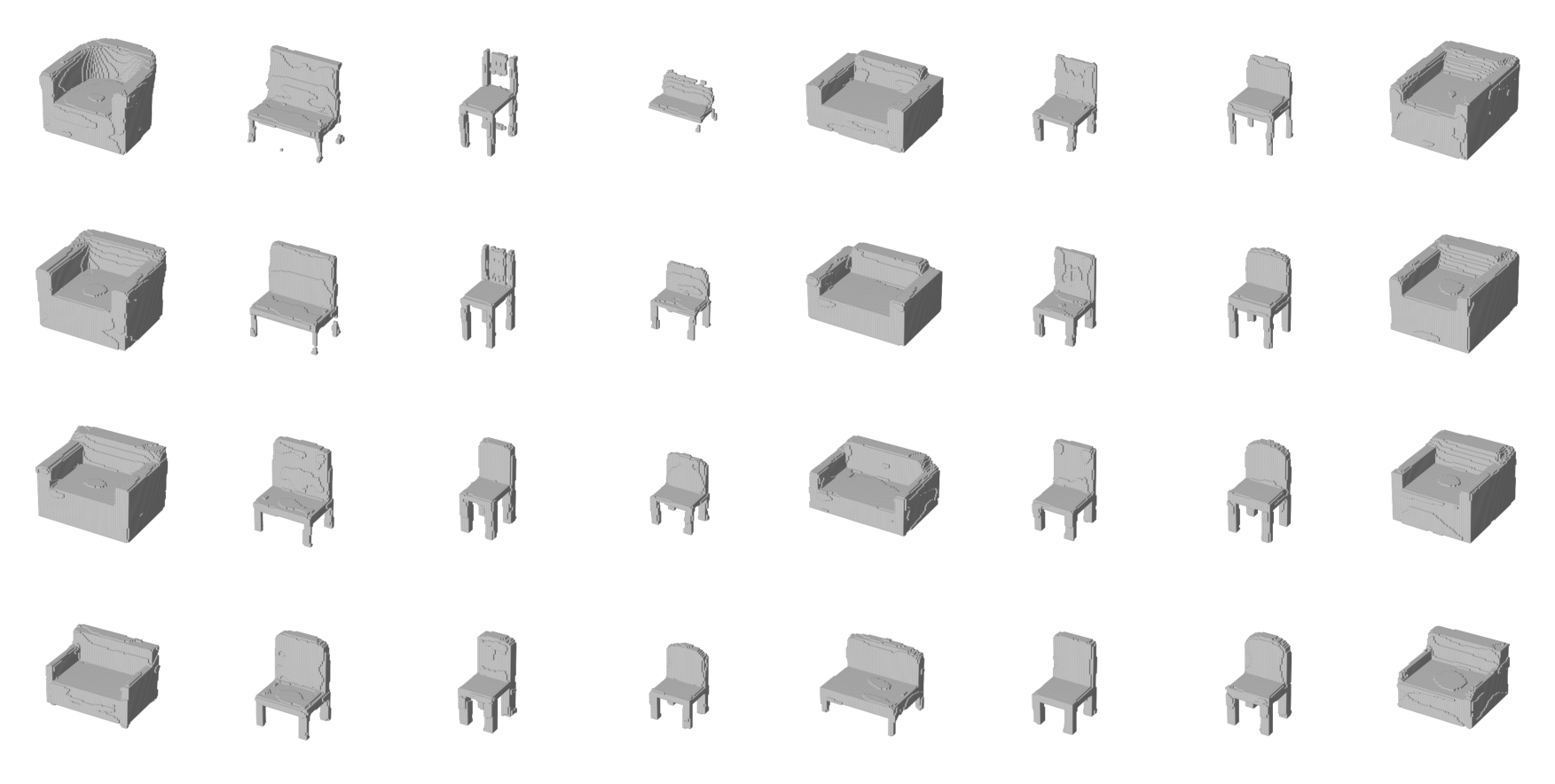}
\end{center}
\vspace{-5mm}
\caption{Uncurated samples from flow trained on 256-dim modulations of ShapeNet Chairs $64^3$, at varying temperatures. Temperatures for each row: 1.0, 0.9, 0.8, 0.7.}
\label{fig:chair-flow-samples-256-vary-temp}
\vspace{-5mm}
\end{figure}

\begin{figure}[t]
\begin{center}
\includegraphics[width=0.8\columnwidth]{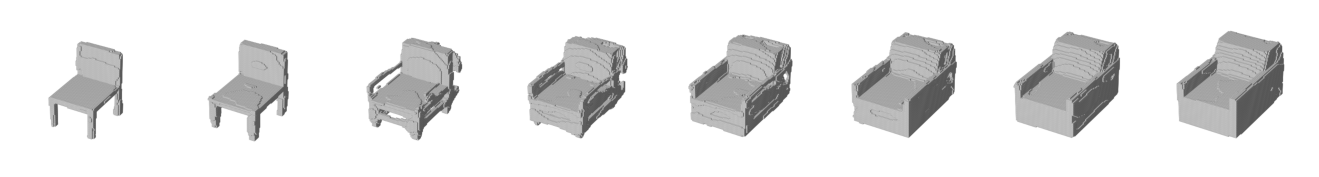}
\includegraphics[width=0.8\columnwidth]{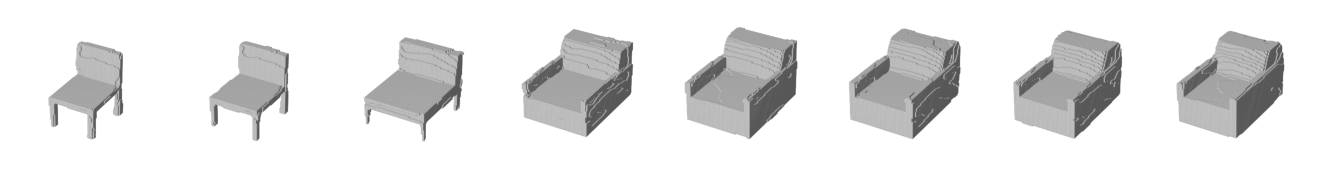}
\end{center}
\vspace{-5mm}
\caption{Comparison of interpolation in modulation space (top) and flow latent space (bottom) for 256-dim modulations of ShapeNet 10 Classes $64^3$.}
\label{fig:chair-latent-interpolation}
%\vspace{-5mm}
\end{figure}

\begin{figure}[h!]
\begin{center}
\includegraphics[width=0.45\columnwidth]{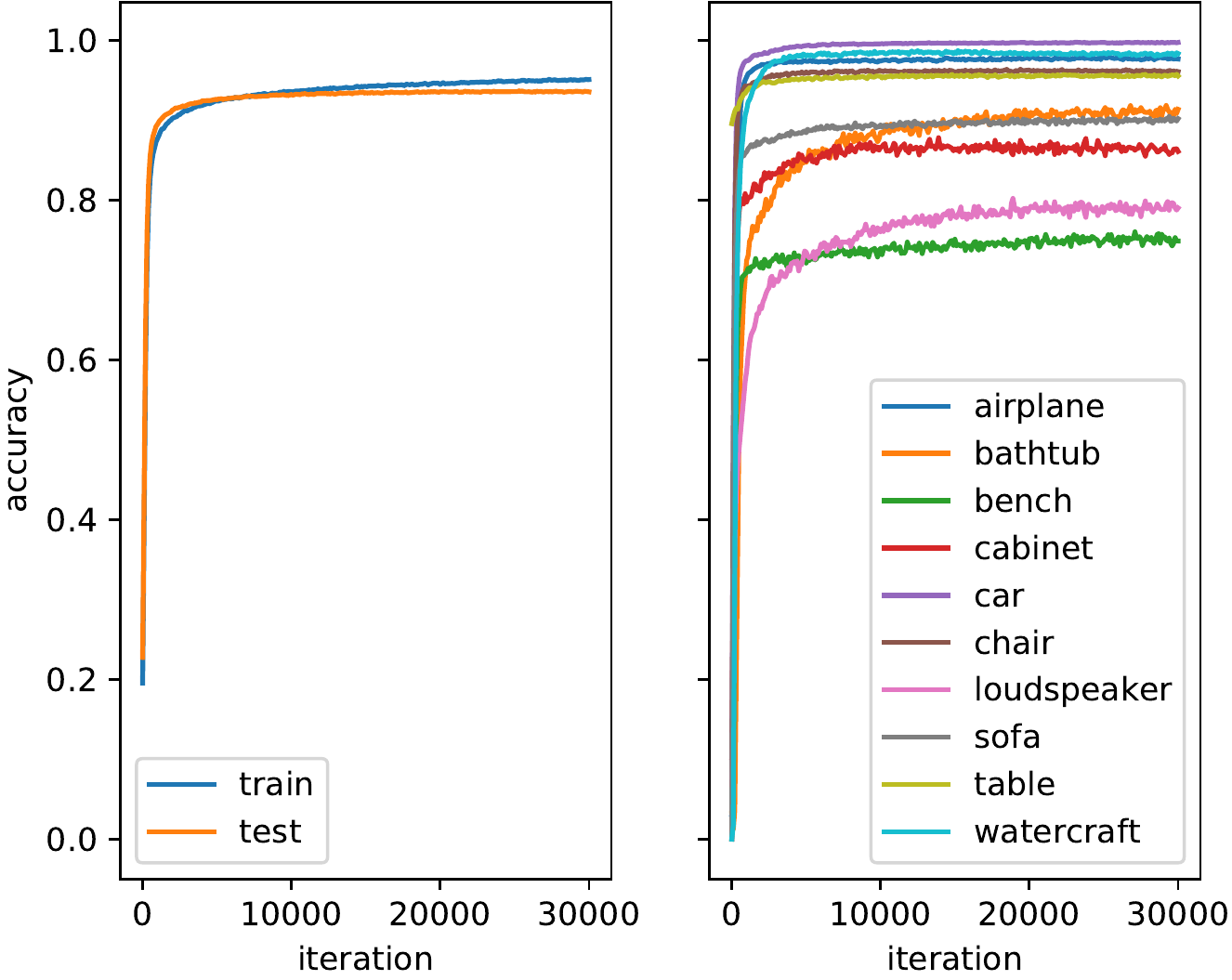}
\unskip\ \vrule\ \includegraphics[width=0.45\columnwidth]{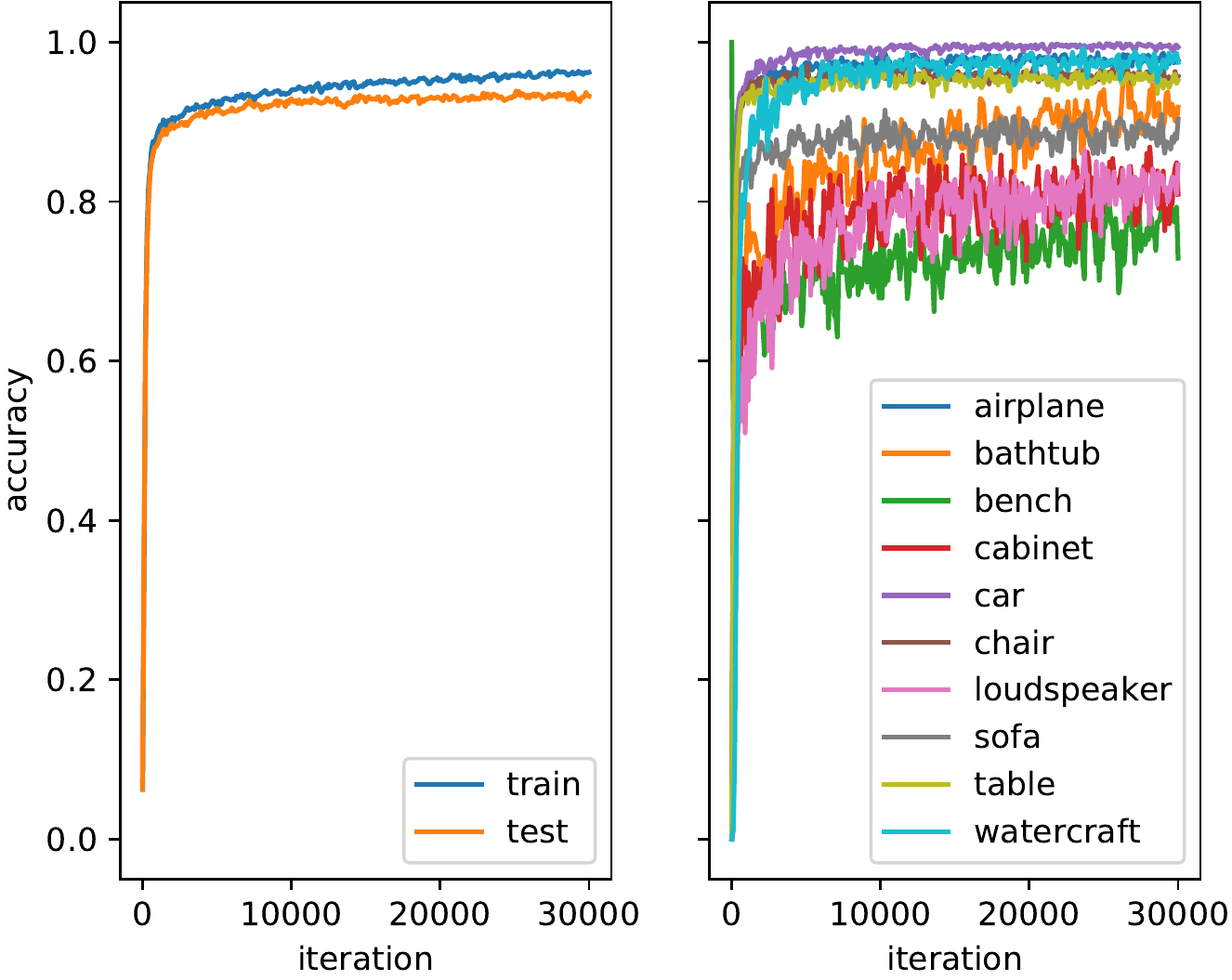}
\end{center}
\caption{Left: Train/test (left) \& per-class test (right) classification accuracy throughout training on 256-dim modulations of ShapeNet 10 Classes $64^3$, using batch size 1024. Right: Same for 3D CNN, but using batch size 64 (GPU memory only allows up to 8 per device). Both Using exponential moving average with decay=0.99 and bs=1024.}
\label{fig:shapenet-classification-training-curve}
\end{figure}

\begin{figure}[h!]
\begin{center}
\includegraphics[width=\columnwidth]{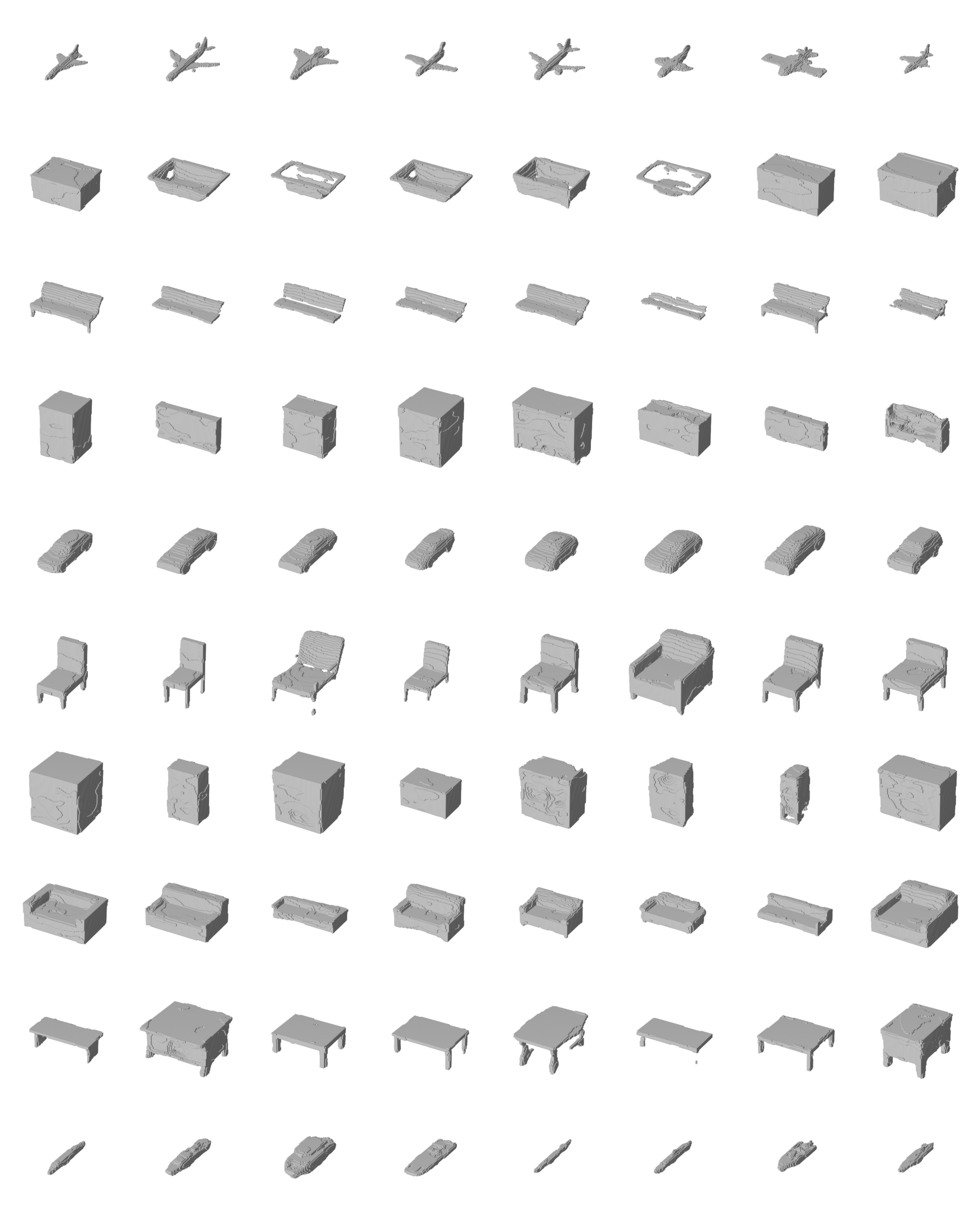}
\end{center}
\caption{Uncurated samples from class-conditional flow trained on 256-dim modulations of ShapeNet 10 classes $64^3$.}
\label{fig:shapenet-top10-class-conditional-flow-samples-256}
\end{figure}

\begin{figure}[h!]
\begin{center}
\includegraphics[width=0.75\columnwidth]{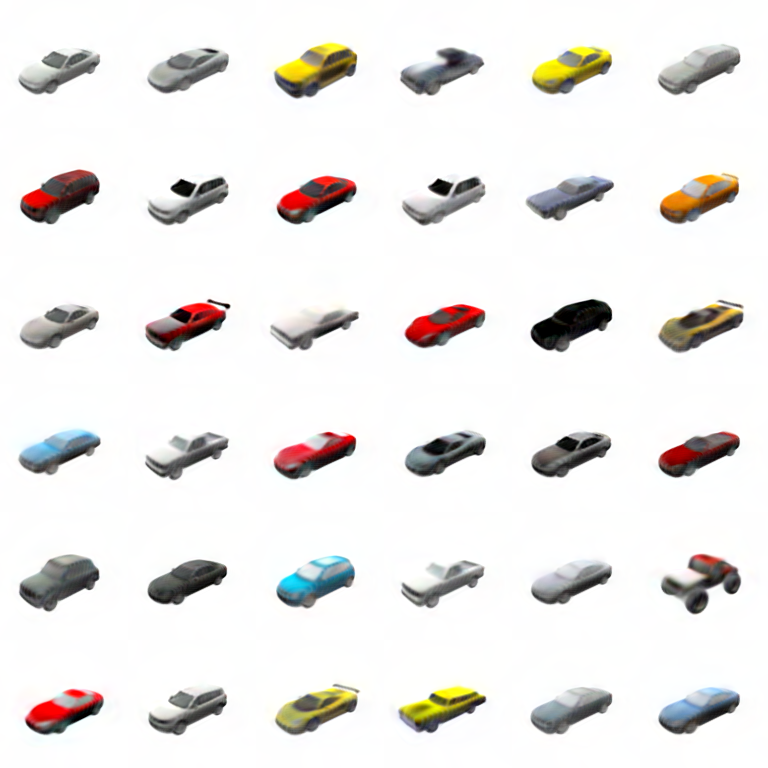}
\end{center}
\caption{Uncurated samples from DDPM trained on 64-dim modulations of SRN Cars.}
\label{fig:car-ddpm-samples}
\end{figure}

\textbf{Uncurated samples from flow on ShapeNet}.
In \autoref{fig:chair-flow-samples-256} we show uncurated samples from the NSF trained on 256-dimensional modulations of ShapeNet Chairs. We also show samples at various temperatures of the base distribution in \autoref{fig:chair-flow-samples-256-vary-temp}, where samples become more globally coherent with lower temperature but show less diversity. Additionally in \autoref{fig:shapenet-top10-class-conditional-flow-samples-256} we show uncurated samples from the class-conditional flow trained on 256-dimensional modulations of ShapeNet 10 classes.

\textbf{ShapeNet 10 classification results for (V)AE latent representations}.
As a baseline, we have fit Autoencoders (AE) and Variational Autoencoders (VAE) \cite{kingma2013auto, rezende2014stochastic} on ShapeNet with 3DCNN enc/decoders and 256-dim latents. Sweeping over hyperparams such as model size gave (V)AEs with reconstruction voxel-acc $\in [98, 99.5]\%$. Classification on resulting (V)AE latents, with the same MLP hyperparam sweep for functa, gave best test accuracy of $93.2\pm0.2\%$ (AE), $91.9\pm0.3\%$ (VAE) - similar if not worse than 256-dim functa. Note that functa provide a unified architectural framework for various modalities, whereas VAEs need modality specific architectures e.g.\ designing a VAE for NeRF scenes is tricky (c.f.\ NeRF-VAE). Also it is unclear how to do inference (imputation, novel view synthesis) from partial observations for VAEs, yet this can be done with functa as shown in the paper.

\textbf{ShapeNet 10 classification on functa for bigger classifiers}. 
In \autoref{tab:classification-further} we show test accuracies and parameter counts for bigger models compared to those shown in \Cref{sec:classification}. Note that bigger models perform better for both classes of models, and bigger 3D CNNs perform better than bigger MLPs on functa, although we need much bigger 3D CNNs to do so. In \autoref{fig:shapenet-classification-training-curve} we also show train/test accuracy and per-class test-accuracy throughout training for both the MLP classifier on modulations and the 3D CNN classifier. The classes that each classifier struggles with are fairly similar.

\begin{table}[h!]
    \centering
    \small
    \begin{sc}
    \begin{tabular}{l|ccc|ccc}
        \toprule[1pt]
        & \multicolumn{3}{c|}{MLP on functa} & \multicolumn{3}{c}{3D CNN on array} \\ \midrule
        Test accuracy & $93.6 \pm 0.1\%$ & $93.9 \pm 0.1\%$ & $94.0 \pm 0.1\%$ & $93.3 \pm 0.3\%$ & $94.5 \pm 0.2\%$ & $94.8 \pm 0.0\%$ \\ \midrule
        $n_{\text{params}}$ & 83k & 212k & 924k & 550k & 2.0M & 7.9M \\
        \bottomrule[1pt]
    \end{tabular}
    \end{sc}
    \caption{Same as \autoref{tab:classification} but for a wider range of model sizes}
    \label{tab:classification-further}
\end{table}

\begin{figure}[t]
\begin{center}
\raisebox{25pt}{\rotatebox[]{90}{Observed}}\hspace{5pt}
\includegraphics[width=0.11\columnwidth]{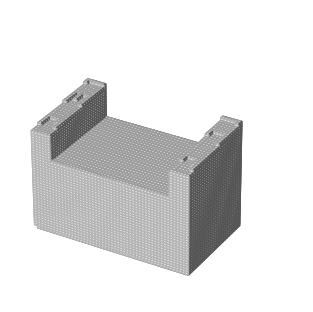}
\includegraphics[width=0.11\columnwidth]{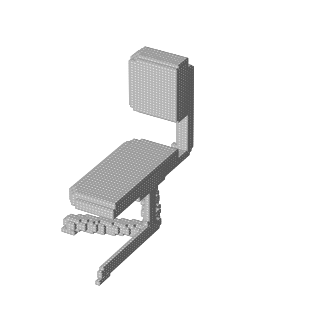}
\includegraphics[width=0.11\columnwidth]{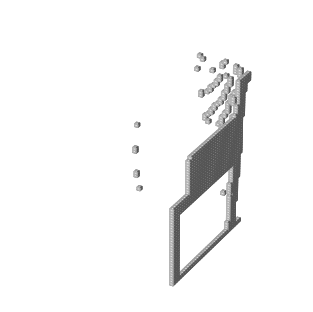}
\includegraphics[width=0.11\columnwidth]{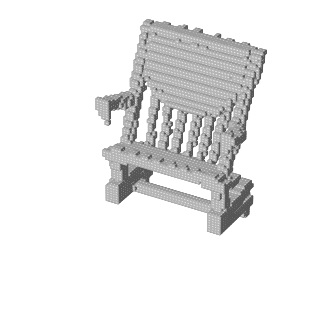}

\raisebox{25pt}{\rotatebox[]{90}{Inferred}}\hspace{5pt}
\includegraphics[width=0.11\columnwidth]{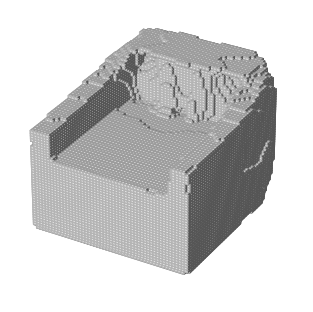}
\includegraphics[width=0.11\columnwidth]{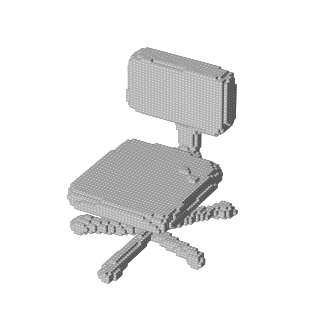}
\includegraphics[width=0.11\columnwidth]{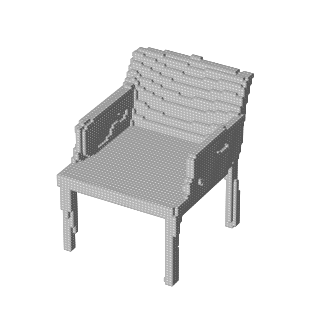}
\includegraphics[width=0.11\columnwidth]{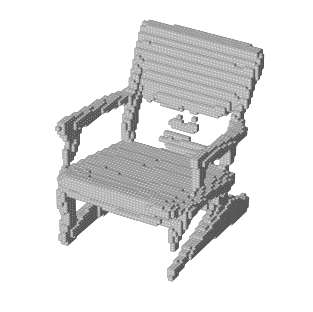}
\end{center}

\vspace{-8pt}
\caption{Additional results to \autoref{fig:chair-imputation} for imputation of different chairs from the test set. GIF showing course of optimization: \href{https://github.com/deepmind/functa\#figure-26}{\nolinkurl{github.com/deepmind/functa\#figure-26}}}
\label{fig:chair-imputation-appendix}
\vspace{-3mm}
\end{figure}

\begin{figure}[t]
\begin{center}
\includegraphics[width=0.47\columnwidth]{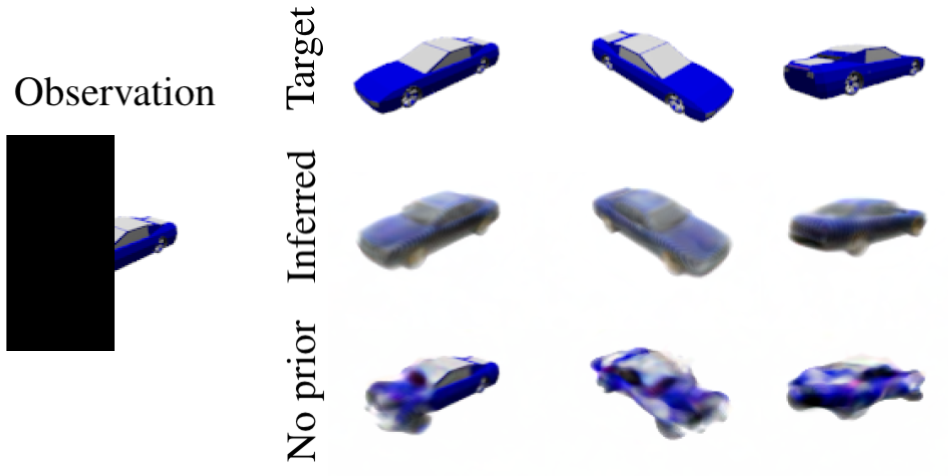}\hspace{0.05\columnwidth}
\includegraphics[width=0.47\columnwidth]{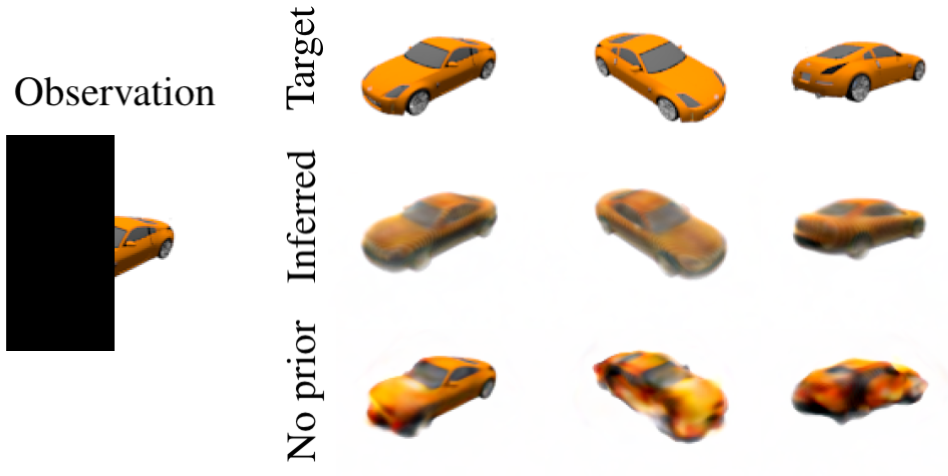}

\vspace{20pt}

\includegraphics[width=0.47\columnwidth]{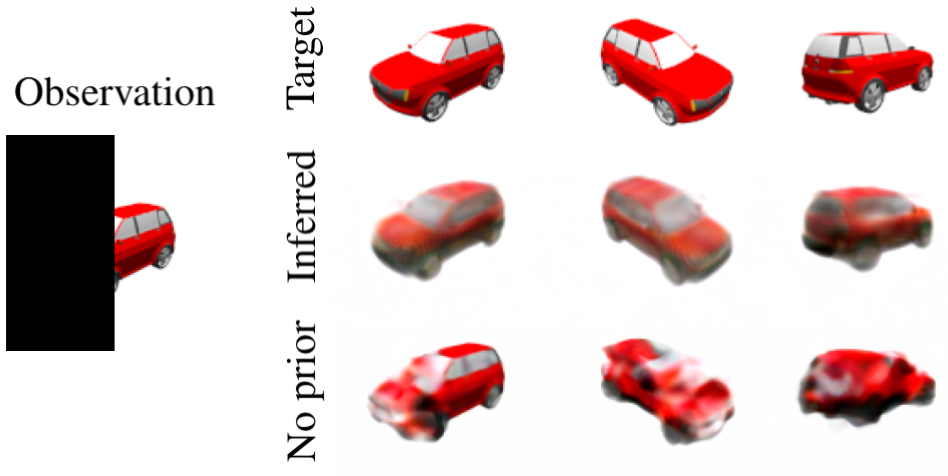}\hspace{0.05\columnwidth}
\includegraphics[width=0.47\columnwidth]{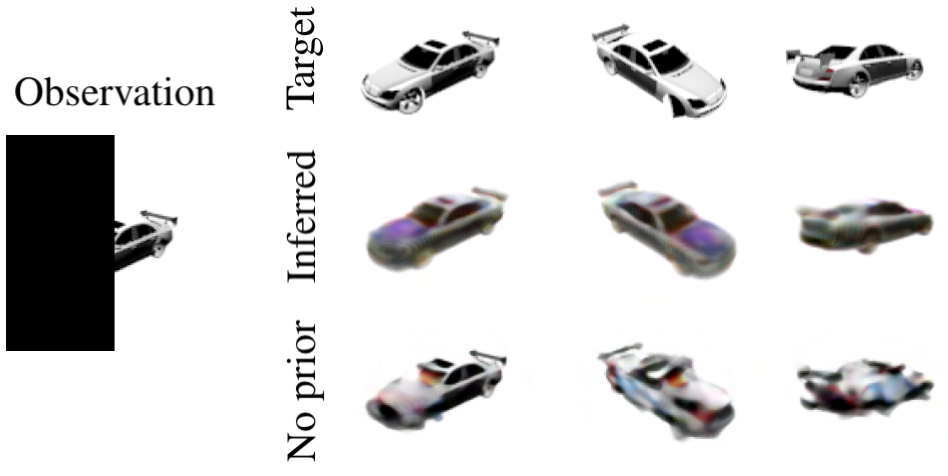}

\end{center}

\vspace{-12pt}
\caption{Additional results to \autoref{fig:novel-view-synthesis} for novel view synthesis from additional occluded test scenes. GIF: \href{https://bit.ly/3x842vO}{\nolinkurl{bit.ly/3x842vO}}}
\label{fig:novel-view-synthesis-appendix}
\vspace{-3mm}
\end{figure}

\end{document}